\documentclass[sigconf]{acmart}

\setcopyright{none}
\renewcommand\footnotetextcopyrightpermission[1]{}
\usepackage{colortbl}       
\usepackage{amsfonts}
\usepackage{nicefrac}
\usepackage{multirow}
\usepackage{pifont}
\usepackage{subcaption}
\usepackage{adjustbox}
\usepackage{enumitem}
\usepackage{wrapfig}
\usepackage{algorithm}
\usepackage{algpseudocode}
\usepackage{float}
\usepackage{titletoc}
\usepackage{nicematrix}
\usepackage{placeins}       

\extrafloats{200}

\definecolor{revision}{rgb}{0.0, 0.0, 0.8}

\definecolor{improve}{rgb}{0.0, 0.5, 0.0}
\definecolor{degrade}{rgb}{0.8, 0.0, 0.0}
\definecolor{firstcolor}{rgb}{0.8, 0.0, 0.0}
\definecolor{secondcolor}{rgb}{0.0, 0.0, 0.8}
\definecolor{darkorange}{rgb}{0.85, 0.40, 0.00}
\definecolor{deepgreen}{rgb}{0.00, 0.55, 0.25}
\definecolor{darkred}{rgb}{0.55, 0.00, 0.00}
\definecolor{darkgreen}{rgb}{0.00, 0.45, 0.20}
\newcommand{\up}[1]{\textcolor{improve}{#1}}

\newcommand{\first}[1]{\textcolor{firstcolor}{\textbf{#1}}}
\newcommand{\second}[1]{\textcolor{secondcolor}{\underline{#1}}}

\title[AdaTKG: Adaptive Memory for Temporal Knowledge Graph Reasoning]{AdaTKG: Adaptive Memory for \\Temporal Knowledge Graph Reasoning}

\author{Seunghan Lee}
\affiliation{%
  \institution{LG AI Research}
  \city{Seoul}
  \country{South Korea}
}
\author{Jun Seo}
\affiliation{%
  \institution{LG AI Research}
  \city{Seoul}
  \country{South Korea}
}
\author{Jaehoon Lee}
\affiliation{%
  \institution{LG AI Research}
  \city{Seoul}
  \country{South Korea}
}
\author{Sungdong Yoo}
\affiliation{%
  \institution{LG AI Research}
  \city{Seoul}
  \country{South Korea}
}
\author{Minjae Kim}
\affiliation{%
  \institution{LG AI Research}
  \city{Seoul}
  \country{South Korea}
}
\author{Tae Yoon Lim}
\affiliation{%
  \institution{LG AI Research}
  \city{Seoul}
  \country{South Korea}
}
\author{Dongwan Kang}
\affiliation{%
  \institution{LG AI Research}
  \city{Seoul}
  \country{South Korea}
}
\author{Hwanil Choi}
\affiliation{%
  \institution{LG AI Research}
  \city{Seoul}
  \country{South Korea}
}
\author{SoonYoung Lee}
\affiliation{%
  \institution{LG AI Research}
  \city{Seoul}
  \country{South Korea}
}
\author{Wonbin Ahn}
\affiliation{%
  \institution{LG AI Research}
  \city{Seoul}
  \country{South Korea}
}

\renewcommand{\shortauthors}{Lee et al.}

\acmConference[KDD '26 GMLLM Workshop]%
  {2nd Frontiers in Graph Machine Learning for the Large Model Era}%
  {August 2026}{Jeju, Korea}

\begin{document}

\begin{abstract}
Temporal knowledge graphs (TKGs) represent time-stamped relational facts and support a wide range of reasoning tasks over evolving events.
However,
existing
methods
produce entity representations that are \textit{static} at the entity level, in that each representation is a function of learned parameters \textit{only} and retains
\textit{no trace of the interactions} in which the entity has participated.
In this paper, we depart from this static view and propose that each entity be modeled as an \textbf{\textit{adaptive process}} whose representation is refined every time the entity participates in a fact.
To this end, we propose \textbf{AdaTKG}, which maintains a per-entity \textit{memory} that is updated with every observed interaction, with the memory accumulating online and predictions improving as more interactions arrive.
Specifically, we instantiate the memory update as a learnable exponential moving average governed by a single \textit{shared} scalar instead of using learnable parameters \textit{for each entity}, enabling AdaTKG to handle entities unseen during training.
Extensive experiments confirm consistent gains over TKG baselines,
demonstrating the effectiveness of adaptive memory.
Code is available at: \url{https://github.com/seunghan96/AdaTKG}.
\end{abstract}

\maketitle

\section{Introduction}
\label{sec:introduction}

Temporal knowledge graphs (TKGs) organize real-world facts into time-stamped relational quadruples and have become a foundation for reasoning about events over time~\cite{tkgsurvey_1}.
They support applications such as event forecasting and risk analysis, each framed as predicting a missing entity or relation at a future timestamp.
A substantial body of work has investigated learning temporal dynamics over such graphs~\cite{REGCN,CENT,Hismatch,LogCL,ECEformer,GenTKG,LLM-DA}, and these models form the backbone of TKG reasoning pipelines.

Despite this rapid progress, essentially every existing TKG reasoning method shares a common design choice that has been left largely unexamined: the representation of any given entity is \textit{static} at the entity level, in the sense that it is a function of the learned parameters alone and \textit{carries no trace of the interactions in which the entity has participated}.
This holds even for the recent line of inductive methods~\cite{ALRE-IR,zrllm,POSTRA,transfir}, which are designed to handle entities unseen during training.

\begin{figure}[t]
\centering
\begin{adjustbox}{max width=0.96\linewidth}
\includegraphics[width=\linewidth]{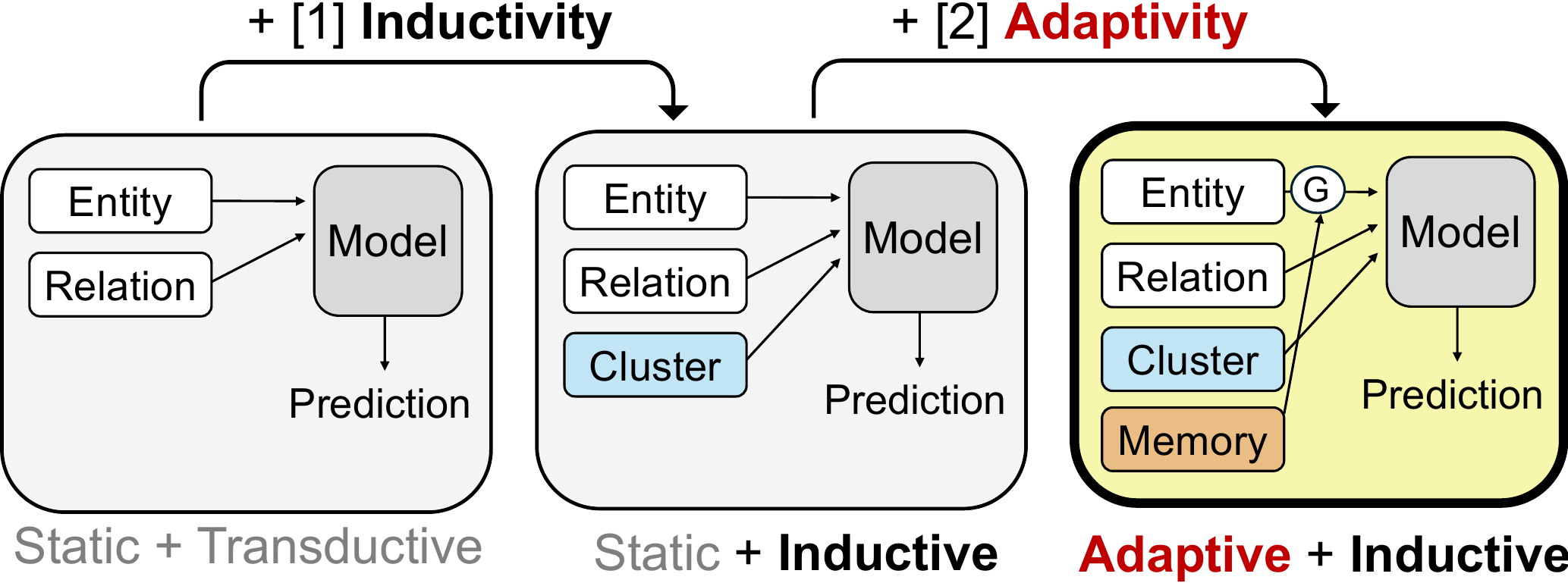}
\end{adjustbox}
\caption{
\textbf{Adaptive TKG reasoning.}
Our framework enables \textit{adaptivity} through
a \textit{memory}
refined with each interaction.
}
\label{fig:intro}
\vspace{-14pt}
\end{figure}
While recent work has focused on whether a TKG reasoner can handle emerging entities unseen at training time (\textit{transductive} vs.\ \textit{inductive}), we focus on a different aspect, 
whether \textit{each entity's representation is refined 
whenever
entity participates in a fact}, 
formalizing
as
\textbf{\textit{static}} vs.\ \textbf{\textit{adaptive}} distinction.
We propose that every entity should be modeled as an \textit{adaptive process} whose representation is refined with each interaction, as illustrated in Figure~\ref{fig:intro}.

To this end, we propose \textbf{AdaTKG}, an \textit{Adaptive Inductive} TKG reasoning method that refines each entity's representation through a \textit{per-entity memory} updated whenever the entity participates in a fact.
Specifically, we instantiate the memory update as a learnable exponential moving average (EMA) that blends 
previous memory with each newly observed interaction via a single learnable scalar.
As the parameter is 
\textit{shared} across entities
and \textit{not assigned per entity},
AdaTKG 
enables \textit{inductive reasoning} on entities unseen during training
while remaining efficient.
Our main contributions are: 
\begin{itemize}[leftmargin=0.6em, itemsep=0.0pt, topsep=0.0pt]
\item We argue that TKG reasoning should be \textit{adaptive} at the entity level, with each entity's representation refined through its own interactions. 
This frames TKG reasoning along a new axis beyond transductive vs. inductive, namely \textit{static} vs.\ \textit{adaptive}.
\item We instantiate this principle with \textbf{AdaTKG}
in which
a learnable EMA blends each entity's past memory with new interactions via a \textit{single shared scalar parameter}. Since no per-entity learnable parameter is introduced, the mechanism is directly applicable to entities unseen during training.
\item We conduct extensive experiments showing that AdaTKG consistently outperforms strong TKG baselines, 
with the gain growing as more interactions 
are observed.
This 
implies that 
\textit{adaptivity} at the entity level
is a crucial component of TKG reasoning.
\end{itemize}

\section{Related Work}
\label{sec:related_work}
\textbf{Reasoning on TKGs.}
Reasoning on temporal knowledge graphs (TKGs) aims to infer missing or future facts by modeling the temporal evolution of entities and relations~\cite{tkgsurvey_1}.
Prior work is typically split into \textit{interpolation}~\cite{TTransE,TNTComplex,TEILP} and \textit{extrapolation}~\cite{CyGNet,REGCN,CENT,Hismatch,LogCL}, with more recent approaches leveraging Transformer architectures and language models~\cite{ECEformer,GenTKG,LLM-DA}.
These methods, however, operate under a closed-world assumption and initialize the embedding of any new entity randomly, leading to representation collapse~\cite{collapse} whenever an entity emerges without historical interactions.

\textbf{Inductive reasoning on KGs.}
Inductive learning on static KGs seeks to generalize to entities
unseen during training, with representative methods exploiting subgraph structure~\cite{Grail,TACT,INDIGO}, meta-learning~\cite{morse}, relation-aware attention~\cite{InGram}, or zero-shot transfer across graphs~\cite{ultra}.
Extending inductivity to the temporal setting, ALRE-IR~\cite{ALRE-IR}, zrLLM~\cite{zrllm}, and POSTRA~\cite{POSTRA} target unseen relations or entities under limited supervision, and TransFIR~\cite{transfir} recently establishes the state of the art by transferring \textit{type-level} behavioral prototypes from the same cluster to each emerging entity.

\textbf{Inductive reasoning without temporal state.}
A complementary line of inductive methods maintains \textit{no temporal per-entity state}.
Static KG foundation models~\cite{Grail,INDIGO,TACT,InGram,ultra} re-derive each entity representation from local graph topology for every query, while PLM- and Transformer-based TKG reasoners~\cite{ICL,PPT,ECEformer} cast reasoning as a text-conditioned or sequence-modeling task whose inductive behavior hinges on an entity's textual surface form.
In both cases, the representation is recomputed per query rather than refined as the entity accumulates interactions.

\textbf{Temporal memory and meta-adaptation.}
Other methods maintain or adapt per-entity state, but not in the forward-only, cold-start inductive regime we target.
Meta-learning approaches~\cite{morse} treat each emerging entity as a meta-task and adapt via gradient updates on a small support set at inference, requiring test-time optimization, whereas dynamic graph neural networks~\cite{dynamic_graph,trivedi2017know,trivedi2019dyrep,xu2020tgat,rossi2020tgn} maintain a time-evolving per-node memory but assume a closed entity set observed at training time.
In contrast, AdaTKG allocates memory at inference and scores an entity from its very first interaction through a forward-only update, with a strict cold-start guarantee and no test-time optimization.

Across the methods above, entity representations are \textit{static at the entity level},
which are a function of the learned parameters alone.
To the best of our knowledge, AdaTKG is the first TKG reasoning method that is \textit{adaptive at the entity level},
maintaining a memory for each entity that is updated with every interaction.

\begin{table}[t]
\centering
\caption{\textbf{Representative formulations on TKGs.}
We argue that the entity should not be \textit{static}, but instead an \textit{adaptive} process whose representation is refined whenever it participates in a fact.
We omit the $(t_q)$ superscript for simplicity.}
\label{tab:paradigms}
\vspace{-6pt}
\begin{adjustbox}{max width=\linewidth}
\small
\begin{tabular}{cc|l}
\toprule
\multicolumn{2}{c}{\textbf{Paradigm}}
&
\multicolumn{1}{|c}{\multirow{2.5}{*}{\textbf{Representation $\mathbf{z}_e$}}} \\
\cmidrule(lr){1-2}
Representation & Generalization & \\
\midrule
\textcolor{secondcolor}{\textbf{Static}} & \textcolor{secondcolor}{\textbf{Transductive}}
& $\mathbf{z}^{\mathrm{ST}}_e = \mathbf{h}_e$ \\
\textcolor{secondcolor}{\textbf{Static}} & \textcolor{firstcolor}{\textbf{Inductive}}
& $\mathbf{z}^{\mathrm{SI}}_e = \mathbf{z}^{\mathrm{ST}}_e \,+\, \textcolor{firstcolor}{\omega_e\cdot\mathbf{c}_{\pi(e)}}$ \\
\textcolor{firstcolor}{\textbf{Adaptive}} & \textcolor{firstcolor}{\textbf{Inductive}}
& $\mathbf{z}^{\mathrm{AI}}_e = \textcolor{firstcolor}{(1\!-\!g_e)}\odot\, \mathbf{z}^{\mathrm{SI}}_e \,+\textcolor{firstcolor}{\, g_e\odot\mathbf{m}_e}$ \\
\bottomrule
\end{tabular}
\end{adjustbox}
\vspace{-8pt}
\end{table}
\section{Problem Definition}
\textbf{Temporal knowledge graph (TKG).}
TKG is a sequence of timestamped snapshots
\begin{equation}
\mathcal{G}=\{\mathcal{G}_t\}_{t\in\mathcal{T}},
\qquad
\mathcal{G}_t=(\mathcal{E}_{1:t},\,\mathcal{R},\,\mathcal{F}_t),
\end{equation}
where $\mathcal{T}$ is the discrete set of timestamps, $\mathcal{E}_{1:t}$ denotes the set of entities observed up to time $t$, $\mathcal{R}$ is the relation set, and
\(
\mathcal{F}_t \subseteq \mathcal{E}_{1:t}\times\mathcal{R}\times\mathcal{E}_{1:t}\times\{t\}
\)
is the set of timestamped facts that hold at time $t$.
Each fact is written as a quadruple $(e_s,\,r,\,e_o,\,t)$, with $e_s,e_o\in\mathcal{E}_{1:t}$ the subject and object entities and $r\in\mathcal{R}$ their relation.
The cumulative history available up to time $t_q$ is
\(
\mathcal{H}_{t_q} \;=\; \bigcup_{i<t_q} \mathcal{F}_i .
\)

\textbf{Link prediction task in TKG.}
Given a query $(e_s,\,r,\,?,\,t_q)$ whose timestamp $t_q$ lies in the future, the TKG link prediction task is to recover the missing entity from the entity pool by ranking every candidate $e\in\mathcal{E}$ under a scoring function $\phi(\cdot)$ conditioned on the history $\mathcal{H}_{t_q}$.
Following the extrapolation protocol, the dataset is split chronologically so that no future information is leaked during training.
As test queries are temporally ordered, the protocol breaks the i.i.d. assumption by design, as earlier ground-truth facts enter $\mathcal{H}_{t_q}$ when scoring later queries.

\textbf{Emerging entities.}
Let the first-appearance time of an entity $e$:
\begin{equation}
t_e(e) \;=\; \min\bigl\{\, t\in\mathcal{T} \,\big|\, e \text{ appears in some fact of } \mathcal{F}_t \,\bigr\}.
\end{equation}
At timestamp $t$, an entity $e$ is an \textit{emerging entity} if $e\in\mathcal{E}_{1:t}\setminus\mathcal{E}_{1:t-1}$, and its participation at $t_q=t_e(e)$ is accompanied by no prior history.
Following previous works~\cite{InGram,transfir},
we adopt this setting as our primary challenge, focusing on 
answering queries at 
the moment an entity enters the graph.
This 
arises frequently in real-world TKGs and
makes reasoning difficult as no interaction history is available for the entity~\cite{tkgsurvey_1}.

\section{From Static TKG to Adaptive TKG}
\label{subsec:paradigms}

In this section, we cast representative prior methods and our proposal into a common scoring form.
Let $\mathbf{h}_e\in\mathbb{R}^d$ denote the \textit{base} representation of entity $e$ and $\mathbf{h}_r\in\mathbb{R}^d$ the embedding of relation $r$.
For a query $(e_q,\,r_q,\,?,\,t_q)$, the score assigned to a candidate $e_o$ takes the unified form
$\phi_{t_q}(e_q,\,r_q,\,e_o) \;=\; f\!\bigl(\,\mathbf{z}^{(t_q)}_{e_q},\; \mathbf{h}_{r_q},\; \mathbf{z}^{(t_q)}_{e_o}\,\bigr)$,
where $f(\cdot)$ is a relational decoder (e.g., ConvTransE~\cite{convtranse}) and $\mathbf{z}^{(t_q)}_{e}$ is the \emph{effective} representation of entity $e$ used for reasoning at time $t_q$.
These formulations differ solely in how $\mathbf{z}^{(t_q)}_e$ is constructed,
with comparison show in Table~\ref{tab:paradigms}.

\begin{figure}[t]
\centering
\begin{adjustbox}{max width=0.96\linewidth}
\includegraphics[width=\linewidth]{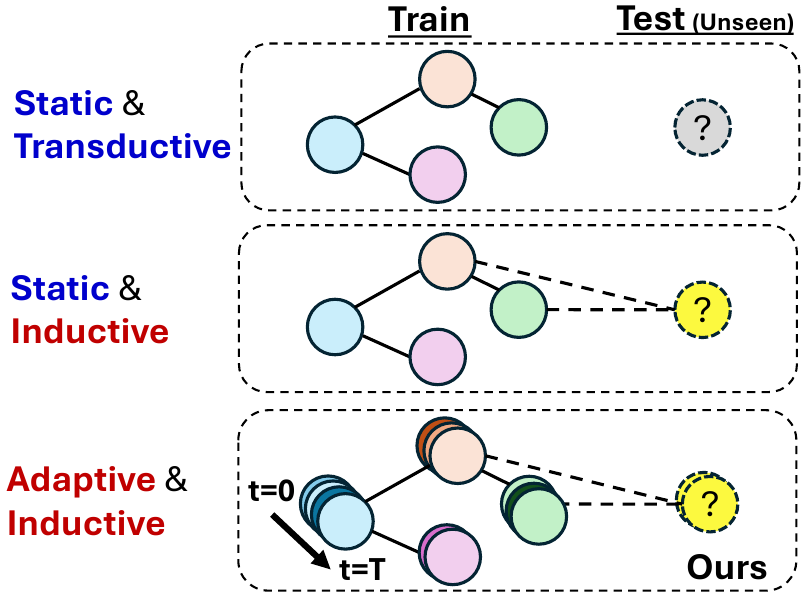}
\end{adjustbox}
\vspace{-3pt}
\caption{
\textbf{Adaptive + Inductive.} 
AdaTKG fuses a 
inductive prior with a per-entity memory,
enabling both generalization to unseen entities and refinement upon each interaction.}
\label{fig:main}
\vspace{-13pt}
\end{figure}

\textbf{Paradigm 1: \textcolor{secondcolor}{Static} + \textcolor{secondcolor}{Transductive}.}
Conventional TKG reasoning methods~\cite{REGCN,CENT,Hismatch,LogCL} learn a lookup embedding for every entity directly from the training facts as
\begin{equation}
\label{eq:static_transductive}
\mathbf{z}^{(t_q)}_e \;=\; \mathbf{h}_e,
\qquad
\mathbf{h}_e \in \mathbb{R}^{d}, \;\; e\in\mathcal{E}_{1:T_{\text{tr}}},
\end{equation}
where $T_{\text{tr}}$ is the final training timestamp.
The representation is \textcolor{black}{\textbf{{static}}} in
that $\mathbf{h}_e$ is a function of the learned parameters alone and 
\textit{carries no trace of the interactions of $e$ itself},
and \textcolor{black}{\textbf{{transductive}}}
in 
that \textit{only entities seen during training} have a well-defined embedding.
However, on emerging entities, 
$\mathbf{h}_e$ is either undefined or initialized randomly, 
leading 
to representation collapse~\cite{collapse}.

\textbf{Paradigm 2: \textcolor{secondcolor}{Static} + \textcolor{firstcolor}{Inductive}.}
To handle emerging entities, recent 
TKG methods~\cite{ALRE-IR,zrllm,POSTRA}
obtain $\mathbf{h}_e$ from an external
source (e.g., a pretrained text encoder) and augment it with a \textit{type-level signal} derived from semantically similar entities.
Let $\pi(e)\in\{1,\ldots,K\}$ be the cluster index of $e$ under a learned codebook and let $\mathbf{c}_{k,\,t}$ denote the 
prototype of cluster $k$ at time $t$, pooled from the interaction patterns of its members~\cite{transfir}. Then, the effective representation is computed using a parametric transfer gate $\Psi(\cdot)$ as
\begin{equation}
\label{eq:static_inductive}
\mathbf{z}^{(t_q)}_e
\;=\;
\mathbf{h}_e \;+\; \omega_e \cdot \mathbf{c}_{\pi(e),\,t_q},
\qquad
\omega_e = \Psi\!\bigl([\mathbf{h}_e\,\|\,\mathbf{c}_{\pi(e),\,t_q}]\bigr).
\end{equation}
This paradigm is \textcolor{black}{\textbf{{inductive}}} in that it \textit{allows an emerging entity to still be scored}, since $\mathbf{c}_{\pi(e),\,t_q}$ is derived from cluster mates rather than from $e$ itself.
However, the transfer is \textcolor{black}{\textbf{{static}}}
in that
$\mathbf{z}^{(t_q)}_e$ is still a function of $\mathbf{h}_e$ and $\mathbf{c}_{\pi(e),\,t_q}$, and 
\textit{does not incorporate any per-entity state beyond them}.

\textbf{Paradigm 3: \textcolor{firstcolor}{Adaptive} + \textcolor{firstcolor}{Inductive}.}
We argue that an entity should be treated not as a \textit{static} representation but as an \textcolor{black}{\textbf{{adaptive}}} process whose representation is refined every time the entity participates in a fact.
To this end, we introduce a \textit{memory} $\mathbf{m}^{(\tau)}_e\in\mathbb{R}^d$ 
for each entity
which is updated via a forward pass whenever $e$ participates in a fact, 
where 
$\tau$ counts the number of updates.
Given a signal $\mathbf{x}^{(\tau)}_e$ summarizing the interaction that triggers the $\tau$-th update of $e$, 
the memory is updated as
\begin{equation}
\label{eq:memory_update}
\mathbf{m}^{(\tau)}_e \;=\; \mathcal{U}\!\bigl(\mathbf{m}^{(\tau-1)}_e,\;\mathbf{x}^{(\tau)}_e\bigr),
\qquad
\mathbf{m}^{(0)}_e \;=\; \mathbf{0},
\end{equation}
where $\mathcal{U}(\cdot,\cdot)$ is a stateful update operator.

At query time $t_q$ we read $\mathbf{m}_e\!\equiv\!\mathbf{m}_e^{(\tau(e,t_q))}$, where $\tau(e,t_q)$ counts the interactions of $e$ strictly before $t_q$.
Note that
following the standard TKG 
protocol~\cite{REGCN,transfir}, training facts are presented in chronological order of $t$.
Memory is thus updated \textit{chronologically}, so no future information leaks into a query.
\textcolor{black}{Concretely, $\mathbf{x}^{(\tau)}e$ is built only from \emph{observed} facts of $e$, so the update never sees the ground-truth object. For an emerging entity $e^*$,
$\mathbf{m}_{e^*}=\mathbf{0}$ and the gate is zero-masked, so AdaTKG reduces exactly to TransFIR (Corollary~1, Appendix~\ref{app:formal_analysis}).}
The effective representation is then produced by an \textit{adaptive gate} $g^{(t_q)}_e\in[0,1]^d$ that fuses the static embedding from Eq.~\eqref{eq:static_inductive} with the 
memory as
\begin{equation}
\label{eq:adaptive_fusion}
\mathbf{z}^{(t_q)}_e
\;=\;
\bigl(1-g^{(t_q)}_e\bigr)\,\odot\,\bigl(\mathbf{h}_e+\omega_e\cdot\mathbf{c}_{\pi(e),\,t_q}\bigr)
\;+\;
g^{(t_q)}_e\,\odot\,\mathbf{m}_e.
\end{equation}
Although the memory is maintained on a per-entity basis and may appear inherently transductive, it can be designed to remain \textcolor{black}{\textbf{{inductive}}}, by governing its update through \textit{entity-agnostic} parameters rather than parameters indexed by individual entities, 
making it applicable to emerging entities.

\begin{figure}[t]
\centering
\begin{adjustbox}{max width=\linewidth}
\includegraphics[width=\linewidth]{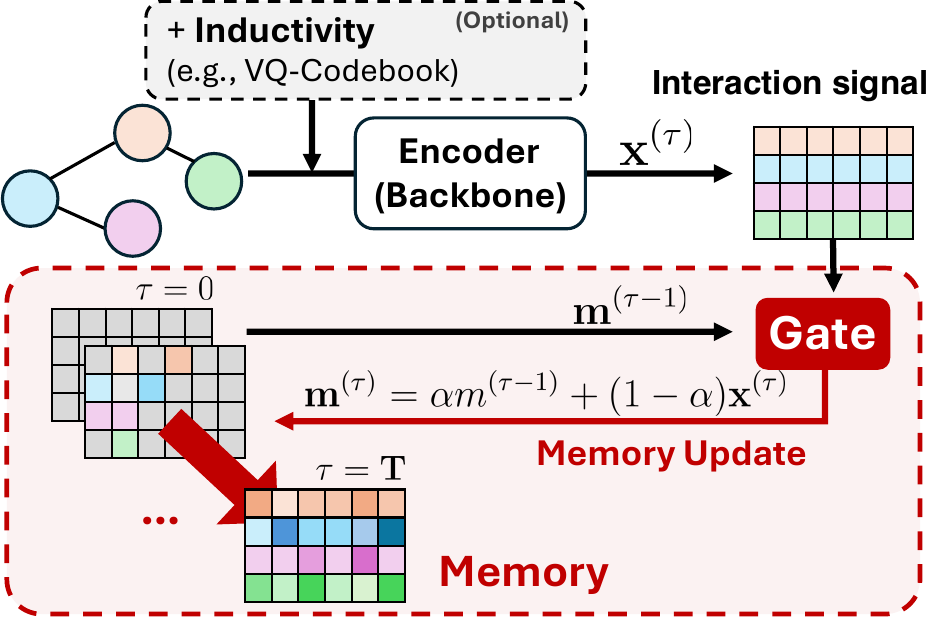}
\end{adjustbox}
\caption{
\textbf{AdaTKG architecture.}
An encoder
produces
an interaction signal $\mathbf{x}^{(\tau)}$ which updates the entity's memory $\mathbf{m}^{(\tau)}$ via a learnable EMA.
The decay rate
$\alpha$
controls how strongly the previous memory is retained relative to the new signal.
}
\label{fig:ema}
\vspace{-13pt}
\end{figure}

\section{Proposed Method: AdaTKG}
\label{sec:adatkg}
\subsection{Model Architecture}
\label{subsec:adatkg}
In this section, we instantiate the components in \textbf{Paradigm 3: \textcolor{firstcolor}{Adaptive} + \textcolor{firstcolor}{Inductive}} in Section~\ref{subsec:paradigms}:
\begin{itemize}[leftmargin=0.6em, itemsep=0.0pt, topsep=0.0pt]
    \item \textbf{[1] Backbone}: Generates static embedding \& interaction signal.
    \item \textbf{[2] Memory update}: Generates memory with interaction signal.
    \item \textbf{[3] Adaptive gate}: Fuses static embedding \& memory.
\end{itemize}

\textbf{[1] Backbone.}
We adopt the backbone of prior work~\cite{transfir}, 
which for each query $q\!=\!(e_q,r_q,?,t)$ generates the three quantities:
\begin{equation*}
\textsc{Backbone}(q) \;\to\; \bigl(\,\mathbf{h}_e,\;\;\omega_e\!\cdot\!\mathbf{c}_{\pi(e),\,t},\;\;\mathbf{x}^{(\tau)}_e\,\bigr),
\end{equation*}
namely \textit{1)} \textit{static entity embedding} $\mathbf{h}_e$ 
from a pretrained text encoder, 
\textit{2)} \textit{type-level inductive prior} $\omega_e\!\cdot\!\mathbf{c}_{\pi(e),\,t}$ from a 
VQ codebook applied to $\mathbf{h}_e$ (Eq.~\eqref{eq:static_inductive}),
and \textit{3)} \textit{interaction signal} $\mathbf{x}^{(\tau)}_e$ produced by a Transformer-based 
encoder over the recent interaction chain of $e$.
Details are provided in Appendix~\ref{app:backbone}.

\textbf{[2] Memory update.}
To make the per-entity memory inductive, it is crucial that 
\textbf{\textit{memory is treated as an internal state, not as a learnable parameter}}.
That is, the buffer $\{\mathbf{m}_e\}_{e\in\mathcal{E}}$ itself carries no trainable weight;
instead, all learnable capacity resides in the \textit{shared} update rule, which is parameterized in an \textit{entity-agnostic} manner.
Consequently, AdaTKG's gain does \textit{not} arise from per-entity knowledge stored in $\mathbf{m}_e$, but from the shared update rule learning to exploit a \textit{new information channel} that opens at inference time and thus remains effective even for emerging entities.

To realize this principle, we instantiate the stateful update operator $\mathcal{U}$ of Eq.~\eqref{eq:memory_update} as a simple yet effective design, a learnable EMA, as shown in Figure~\ref{fig:ema}:
\begin{equation}
\label{eq:ema_update}
\mathbf{m}^{(\tau)}_e \;=\; \alpha\, \mathbf{m}^{(\tau-1)}_e \;+\; (1-\alpha)\, \mathbf{x}^{(\tau)}_e,
\end{equation}
with
$\mathbf{m}^{(0)}_e = \mathbf{0}$,
where $\alpha \;=\; \sigma(\rho)$, $\rho\!\in\!\mathbb{R}$ is a single learnable scalar shared across all entities, and $\sigma(\cdot)$ is the sigmoid, yielding a decay rate $\alpha\!\in\!(0,1)$.
Crucially, $\rho$ is the only learnable parameter,
and it is entity-agnostic by construction, so the same update rule applies to every entity including emerging entities.
Note that EMA is merely one of possible instantiations of $\mathcal{U}$, and \textbf{\textit{our focus is not on the EMA operator itself but on endowing each entity with adaptivity through a per-entity memory}}.
Details of other possible instantiations are provided in Appendix~\ref{app:operators}.

Since the memory is updated online over the training stream, three protocol decisions are required to keep the update strictly causal,
with details provided in Appendix~\ref{app:formal_analysis}.
\begin{itemize}[leftmargin=0.6em, itemsep=0.0pt, topsep=0.0pt]
  \item \textbf{a) Epoch reset.} Buffer is reset at every epoch start, so $\rho,\mathbf{W}_g$ receive a consistent learning signal.
  \item \textbf{b) Chronological replay.} Facts are replayed in time order, so each query reads EMA of past facts.
  \item \textbf{c) Detached update.} With $\mathbf{m}^{(\tau-1)}_e$ detached before the update, no recurrence gradients flow.
\end{itemize}
\textbf{[3] Adaptive gate.}
The 
gate $g^{(t_q)}_e\!\in\![0,1]^d$ in Eq.~\eqref{eq:adaptive_fusion} fuses the 
\textit{static} entity embedding
with the 
\textit{adaptive} 
memory.
Note that $g^{(t_q)}_e$ is a $d$-dimensional gate,
so the fusion in Eq.~\eqref{eq:adaptive_fusion} is a coordinate-wise interpolation and the (cold-start) zero-mask $g^{(t_q)}_e\!=\!\mathbf{0}$ refers to the $d$-dimensional zero vector.
We parameterize it by a sigmoid-gated linear projection with explicit zero-masking when the memory buffer is empty,
where $\mathbf{W}_g\!\in\!\mathbb{R}^{d\times 2d}$ is a learnable projection,
\begin{equation}
\label{eq:gate}
g^{(t_q)}_e \;=\;
\begin{cases}
\mathbf{0} & \text{if } \mathbf{m}_e = \mathbf{0},\\[2pt]
\sigma\!\bigl(\mathbf{W}_g\,[\mathbf{h}_e \,\|\, \mathbf{m}_e]\bigr) & \text{otherwise}.
\end{cases}
\end{equation}
The zero-masking
realizes the hypothesis of Corollary~1 in Appendix~\ref{app:formal_analysis}, guaranteeing that AdaTKG reduces to the Static + Inductive paradigm (Eq.~\eqref{eq:static_inductive}) whenever no interaction of $e$ has been observed.
For the loss function, we follow that of the previous work~\cite{transfir}.


\makeatletter
\setlength{\@fptop}{-55pt}   
\setlength{\@fpsep}{4pt}
\setlength{\@fpbot}{0pt}
\makeatother
\setlength{\textfloatsep}{4pt}

\begin{algorithm}[!t]
\caption{Training of AdaTKG (Single-epoch view). 
}
\fontsize{8.5pt}{9pt}\selectfont  
\label{alg:adatkg_train_one_epoch}
\begin{algorithmic}[1]
\Require Facts $\mathcal{F}_\text{tr}$; Parameters $\rho,\mathbf{W}_g$ (+ backbone $\Theta_\text{bb}$); Per-entity state $\{(\mathbf{m}_e,\tau_e)\}_{e\in\mathcal{E}}$.
\State $\alpha \gets \sigma(\rho)$ and $\mathbf{m}_e \gets \mathbf{0},\ \tau_{e}\gets 0$ \ \textbf{for all} $e\in\mathcal{E}$
\For{timestamp $t = 1$ \textbf{to} $T_{\mathrm{tr}}$}
  \For{each query $q=(e_q,r_q,?,t)$ from the facts at $t$}
    \Statex \quad $\triangleright$ \textcolor{black}{\textit{[1] Embeddings (w/ inductivity)}}
    \State $\bigl(\mathbf{h}_{e_q},\;\omega_{e_q}\!\cdot\!\mathbf{c}_{\pi(e_q),t},\;\mathbf{x}^{(\tau_{e_q}+1)}_{e_q}\bigr) \!\gets\! \textsc{Backbone}(q)$
    \State $\widetilde{\mathbf{h}}_{e} =\bigl(\mathbf{h}_{e}\!+\!\omega_{e}\!\cdot\!\mathbf{c}_{\pi(e),t}\bigr)$
    \Statex \quad $\triangleright$ \textcolor{darkred}{\textit{[2] Memory update}}
    \State $\widetilde{\mathbf{m}}_{e_q} \!\gets\! \mathrm{stop\_grad}(\mathbf{m}_{e_q})$
    \State \textcolor{darkred}{$\mathbf{m}_{e_q} \!\gets\! \alpha\,\widetilde{\mathbf{m}}_{e_q} \;+\; (1\!-\!\alpha)\,\mathbf{x}^{(\tau_{e_q}+1)}_{e_q}$} \Comment{Eq.~\eqref{eq:ema_update}}
    \Statex \quad $\triangleright$ \textcolor{darkgreen}{\textit{[3] Adaptive-gate fusion}}
    \State \textcolor{darkgreen}{$\mathbf{z}^{(t)}_{e} \!\gets\! \bigl(1-g^{(t)}_{e}\bigr)\!\odot\! \widetilde{\mathbf{h}}_{e}  + g^{(t)}_{e}\!\odot\!\mathbf{m}_{e}$} for $e\!\in\!\{e_q\}\cup\mathcal{E}$ \Comment{Eq.~\eqref{eq:adaptive_fusion}}
    \State Score $\phi(e_q,r_q,e_o)\!=\!f(\mathbf{z}^{(t)}_{e_q},\mathbf{h}_{r_q},\mathbf{z}^{(t)}_{e_o})$
    \State $\tau_{e_q} \!\gets\! \tau_{e_q} + 1$
  \EndFor
\EndFor
\end{algorithmic}
\end{algorithm}

\subsection{Properties of the Adaptive Memory}
\label{subsec:cold_to_warm}
\textbf{[Generality] Update-operator agnostic.}
The adaptive memory framework is not tied to EMA, as any stateful update operator $\mathcal{U}$ that maintains a per-entity state can be plugged in without changing the 
training protocol.
It is important to note that our focus is \textit{not on the choice of EMA itself} but on the \textit{adaptive property} it enables; 
to highlight this, we additionally instantiate $\mathcal{U}$ as a GRU cell and as a cross-attention readout over a bounded per-entity buffer,
with details 
provided in Appendix~\ref{app:operators}.

\textbf{[Efficiency] Shared parameters.}
\textcolor{black}{AdaTKG introduces two learnable components:
(i) the EMA module, which contributes a 
shared scalar $\rho$ that governs the memory dynamics, and (ii) the adaptive gate, which contributes a 
shared projection $\mathbf{W}_g$ that controls the fusion. Crucially, neither $\rho$ nor $\mathbf{W}_g$ is indexed by the entity, so the buffer $\{\mathbf{m}_e\}$ 
carries no learnable parameter.}
AdaTKG thus applies uniformly to every entity, remaining effective on
unseen entities.

\textbf{[Online update] Accumulation at inference.}
The per-entity memory $\mathbf{m}_e$ accumulates online with every new fact at inference time, 
so the predictions improve as more interactions 
arrive. 
This online dynamic also underlies the consistent gain on emerging/unknown entities as shown in Table~\ref{tab:sota_comparison},
where the $\mathbf{h}_e$ provides no entity-specific signal and $\mathbf{m}_e$ supplies a substantial new source of information.
Note that $\mathbf{m}_e$ is carried chronologically across 
train/valid/test phases 
(resetting only at training-epoch boundaries), so the memory accumulated during training remains available at test time.

\textbf{[Interpretability] Transition across regimes.}
The adaptive fusion
behaves predictably across the cold-start and warm regimes. With no interaction of $e$ observed, we enforce $g^{(t_q)}_e\!=\!\mathbf{0}$ via a zero-mask on the gate, so that Eq.~\eqref{eq:adaptive_fusion} recovers the \textit{Static + Inductive} formulation in Eq.~\eqref{eq:static_inductive}. As more interactions
are observed, $\mathbf{m}_e$ grows informative and the gate shifts weight toward the memory branch. 

\begin{table*}[t]
\centering
\caption{
\textbf{Comparison with the TKG reasoning methods across emerging entities.}
The table shows the link-prediction performance 
for queries whose query entity first appears at the query timestamp.
Best scores are in \first{bold red}, second-best in \second{underlined blue}.
}
\label{tab:main_results}
\adjustbox{max width=\linewidth}{
\footnotesize
\begin{tabular}{l|ccc|ccc|ccc|ccc}
\toprule
& \multicolumn{3}{c|}{\textbf{ICEWS14}}
& \multicolumn{3}{c|}{\textbf{ICEWS18}}
& \multicolumn{3}{c|}{\textbf{ICEWS05-15}}
& \multicolumn{3}{c}{\textbf{GDELT}} \\
\cmidrule(lr){2-4} \cmidrule(lr){5-7} \cmidrule(lr){8-10} \cmidrule(lr){11-13}
& MRR & H@3 & H@10
& MRR & H@3 & H@10
& MRR & H@3 & H@10
& MRR & H@3 & H@10 \\
\midrule
\multicolumn{13}{l}{\textit{(i) Graph-based}} \\
\midrule
CyGNet~\cite{CyGNet}      & .0111 & .0098 & .0202 & .0031 & .0020 & .0047 & .0031 & .0020 & .0048 & .0067 & .0031 & .0147 \\
RE-GCN~\cite{REGCN}       & .1175 & .1263 & .2232 & .0947 & .1004 & .1797 & .0887 & .0961 & .1803 & .0222 & .0209 & .0393 \\
HiSMatch~\cite{Hismatch}  & .0284 & .0285 & .0418 & .0055 & .0058 & .0076 & .0242 & .0238 & .0443 & .0159 & .0141 & .0270 \\
MGESL~\cite{mgesl}        & .0309 & .0361 & .0603 & .0747 & .0809 & .1031 & .1069 & .1166 & .1563 & .0516 & .0471 & .0815 \\
LogCL~\cite{LogCL}        & .1354 & .1770 & .2273 & .0903 & .1064 & .1548 & .1917 & .2452 & .2855 & .0473 & .0479 & .0973 \\
HisRes~\cite{HisRes}      & .1169 & .1107 & .2132 & .0445 & .0434 & .0735 & .1325 & .1332 & .1407 & .0416 & .0737 & .0932 \\
MLEMKD~\cite{MLEMKD}      & .0685 & .0728 & .1303 & .0402 & .0382 & .0831 & .0833 & .0848 & .1717 & .0229 & .0215 & .0436 \\
\midrule
\multicolumn{13}{l}{\textit{(ii) Path-based}} \\
\midrule
TLogic~\cite{TLogic}      & .0122 & .0107 & .0257 & .0141 & .0131 & .0262 & .0121 & .0108 & .0285 & .0733 & .0749 & .1131 \\
TILP~\cite{Tilp}          & .0397 & .0471 & .1114 & .0498 & .0669 & .1659 & .0358 & .0374 & .1030 & .0053 & .0025 & .0084 \\
ECEformer~\cite{ECEformer}& .0461 & .0496 & .0915 & .0323 & .0680 & .0454 & .0587 & .0642 & .1141 & .0426 & .0410 & .0872 \\
GenTKG~\cite{GenTKG}      & ---    & .0983 & .1919 & ---    & .0540 & .1512 & ---    & .1105 & .1873 & ---    & .0734 & .1013 \\
\midrule
\multicolumn{13}{l}{\textit{(iii) Static inductive}} \\
\midrule
CompGCN~\cite{CompGCN}    & .0682 & .0906 & .1213 & .0638 & .0745 & .1049 & .1885 & .2103 & .2479 & .0472 & .0791 & .0934 \\
ICL~\cite{ICL}            & .0252 & .0261 & .0388 & .0639 & .0727 & .0938 & .0254 & .0302 & .0373 & .0277 & .0326 & .0362 \\
PPT~\cite{PPT}            & .0093 & .1062 & .1716 & .0368 & .0386 & .0650 & .0015 & .0005 & .0022 & .0406 & .0425 & .0764 \\
MorsE~\cite{morse}        & .0136 & .0074 & .0185 & .0078 & .0075 & .0126 & .0381 & .0167 & .0439 & .0039 & .0040 & .0152 \\
InGram~\cite{InGram}      & .0563 & .0596 & .1138 & .0254 & .0265 & .0518 & .0771 & .0793 & .1454 & .0471 & .0430 & .0847 \\
TransFIR~\cite{transfir}  & \second{.1763} & \second{.2096} & \second{.3413}
                          & \second{.1114} & \second{.1230} & \second{.2252}
                          & \second{.2177} & \second{.2530} & \second{.3708}
                          & \second{.1013} & \second{.0994} & \second{.2131}  \\
\midrule
\cellcolor{yellow!15}\textbf{AdaTKG (Ours)}
& \cellcolor{yellow!15}\first{.2011}
& \cellcolor{yellow!15}\first{.2250}
& \cellcolor{yellow!15}\first{.3621}
& \cellcolor{yellow!15}\first{.1379}
& \cellcolor{yellow!15}\first{.1543}
& \cellcolor{yellow!15}\first{.2612}
& \cellcolor{yellow!15}\first{.2270}
& \cellcolor{yellow!15}\first{.2573}
& \cellcolor{yellow!15}\first{.3850}
& \cellcolor{yellow!15}\first{.1051}
& \cellcolor{yellow!15}\first{.1129}
& \cellcolor{yellow!15}\first{.2301} \\
\textit{$\Delta$ over SoTA (\%)}
                          & \textbf{\up{+14.1}} & \textbf{\up{+7.3}}  & \textbf{\up{+6.1}}
                          & \textbf{\up{+23.8}} & \textbf{\up{+25.4}} & \textbf{\up{+16.0}}
                          & \textbf{\up{+4.3}}  & \textbf{\up{+1.7}}  & \textbf{\up{+3.8}}
                          & \textbf{\up{+3.8}}  & \textbf{\up{+13.6}} & \textbf{\up{+8.0}}     \\
\bottomrule
\end{tabular}
}
\end{table*}

\begin{table*}[t]
\centering
\caption{
\textbf{Comparison with the SoTA baseline across the emerging/unknown entities.}
The table shows the link-prediction performance
across the two test slices: \textit{Emerging} and \textit{Unknown}.
}
\label{tab:sota_comparison}
\adjustbox{max width=\linewidth}{
\footnotesize
\begin{tabular}{c|l|ccc|ccc|ccc|ccc}
\toprule
\multirow{2.5}{*}{\textbf{Entity}} & \multicolumn{1}{c|}{\multirow{2.5}{*}{\textbf{Method}}}
& \multicolumn{3}{c|}{\textbf{ICEWS14}}
& \multicolumn{3}{c|}{\textbf{ICEWS18}}
& \multicolumn{3}{c|}{\textbf{ICEWS05-15}}
& \multicolumn{3}{c}{\textbf{GDELT}} \\
\cmidrule(lr){3-5}\cmidrule(lr){6-8}\cmidrule(lr){9-11}\cmidrule(lr){12-14}
& & MRR & H@3 & H@10
  & MRR & H@3 & H@10
  & MRR & H@3 & H@10
  & MRR & H@3 & H@10 \\
\midrule
\multirow{3.5}{*}{\textit{Emerging}}
 & TransFIR
   & .1763 & .2096 & .3413
   & .1114 & .1230 & .2252
   & .2177 & .2530 & .3708
   & .1013 & .0994 &        .2131  \\

 & \cellcolor{yellow!15}\textbf{AdaTKG}
& \cellcolor{yellow!15}\first{.2011}
& \cellcolor{yellow!15}\first{.2250}
& \cellcolor{yellow!15}\first{.3621}
& \cellcolor{yellow!15}\first{.1379}
& \cellcolor{yellow!15}\first{.1543}
& \cellcolor{yellow!15}\first{.2612}
& \cellcolor{yellow!15}\first{.2270}
& \cellcolor{yellow!15}\first{.2573}
& \cellcolor{yellow!15}\first{.3850}
& \cellcolor{yellow!15}\first{.1051}
& \cellcolor{yellow!15}\first{.1129}
& \cellcolor{yellow!15}\first{.2301} \\
\cmidrule(lr){2-14}
 & \textit{$\Delta$ (\%)}
   & \textbf{\up{+14.1}} & \textbf{\up{+7.3}}  & \textbf{\up{+6.1}}
   & \textbf{\up{+23.8}} & \textbf{\up{+25.4}} & \textbf{\up{+16.0}}
   & \textbf{\up{+4.3}}  & \textbf{\up{+1.7}}  & \textbf{\up{+3.8}}
   & \textbf{\up{+3.8}}  & \textbf{\up{+13.6}} & \textbf{\up{+8.0}}    \\
\midrule
\multirow{3.5}{*}{\textit{Unknown}}
 & TransFIR
   & .2136 & .2483 & .3901
   & .1348 & .1517 & .2591
   & .3013 & .3425 & .4706
   & .1080 & .1140 & .2261 \\
 & \cellcolor{yellow!15}\textbf{AdaTKG}
& \cellcolor{yellow!15}\first{.2293}
& \cellcolor{yellow!15}\first{.2590}
& \cellcolor{yellow!15}\first{.4029}
& \cellcolor{yellow!15}\first{.1687}
& \cellcolor{yellow!15}\first{.1897}
& \cellcolor{yellow!15}\first{.3027}
& \cellcolor{yellow!15}\first{.3044}
& \cellcolor{yellow!15}\first{.3458}
& \cellcolor{yellow!15}\first{.4790}
& \cellcolor{yellow!15}\first{.1148}
& \cellcolor{yellow!15}\first{.1208}
& \cellcolor{yellow!15}\first{.2343} \\
\cmidrule(lr){2-14}
 & \textit{$\Delta$ (\%)}
   & \textbf{\up{+7.4}}  & \textbf{\up{+4.3}}  & \textbf{\up{+3.3}}
   & \textbf{\up{+25.1}} & \textbf{\up{+25.1}} & \textbf{\up{+16.8}}
   & \textbf{\up{+1.0}}  & \textbf{\up{+1.0}}  & \textbf{\up{+1.8}}
   & \textbf{\up{+6.3}}  & \textbf{\up{+6.0}}  & \textbf{\up{+3.6}}  \\
\bottomrule
\end{tabular}
}
\end{table*}

\section{Experiments}
\label{sec:experiments}
\textbf{Datasets.}
We evaluate AdaTKG on four widely used temporal knowledge graph benchmarks: ICEWS14, ICEWS18, ICEWS05-15~\cite{DVN_28075_2015}, and GDELT~\cite{leetaru2013gdelt}.
Each dataset is a chronologically ordered sequence of timestamped quadruples $(e_s,\,r,\,e_o,\,t)$.
Details of datasets are discussed in Appendix~\ref{app:dataset}.

\textbf{Evaluation protocol.}
Following prior work on inductive TKG reasoning~\cite{transfir,ALRE-IR}, we evaluate link prediction 
using Mean Reciprocal Rank (MRR) and Hits@$k$ with $k\!\in\!\{3,10\}$, computed over the filtered ranking of every candidate entity.
While we focus on \textit{emerging} entities following previous works~\cite{transfir,InGram}, we additionally evaluate on the \textit{unknown} slice to capture entities that, although unseen in training, may accumulate a few interactions before the query timestamp,
defined as:
\begin{itemize}[leftmargin=0.6em, itemsep=0.0pt, topsep=0.0pt]
\item \textit{Emerging}: Queries on entities that appear for the \textit{very first time} at the query timestamp.
\item \textit{Unknown}: Queries on entities that were never observed during training but \textit{may have appeared in validation or earlier in the test stream}, i.e., $s \in \mathcal{V}\!\setminus\!\mathcal{V}_{\mathrm{train}}$ or $o \in \mathcal{V}\!\setminus\!\mathcal{V}_{\mathrm{train}}$.
\end{itemize}
Note that these two slices are not disjoint: every \textit{Emerging} query is also \textit{Unknown}.
We follow the standard chronological split to prevent leakage of future information into training.

\textbf{Baselines.}
We adopt the same baseline set as TransFIR~\cite{transfir}, with methods grouped into three categories.
\textit{(i) Graph-based}: CyGNet~\cite{CyGNet}, RE-GCN~\cite{REGCN}, HiSMatch~\cite{Hismatch}, MGESL~\cite{mgesl}, LogCL~\cite{LogCL}, HisRes~\cite{HisRes}, and MLEMKD~\cite{MLEMKD}.
\textit{(ii) Path-based}: TLogic~\cite{TLogic}, TILP~\cite{Tilp}, ECEformer~\cite{ECEformer}, and GenTKG~\cite{GenTKG}.
\textit{(iii) Static inductive}: CompGCN~\cite{CompGCN}, ICL~\cite{ICL}, PPT~\cite{PPT}, MorsE~\cite{morse}, InGram~\cite{InGram}, and TransFIR~\cite{transfir}.
All baseline numbers are taken from TransFIR~\cite{transfir}, with TransFIR itself reproduced from its official repository for fair comparison.

\textbf{Implementation details.}
AdaTKG initializes 
$\mathbf{h}_e$ from a frozen BERT~\cite{bert} encoder applied to the entity's textual surface form.
Relational scoring is performed with ConvTransE~\cite{convtranse}, the decoder adopted by recent TKG baselines.
We 
use Adam optimizer~\cite{kingma2014adam}
and select hyperparameters via validation MRR with early stopping.
Results averaged over three random seeds, along with standard deviations, are 
in Appendix~\ref{app:multiseed}.
Further details are 
in Appendix~\ref{app:implementation}.

\subsection{Main Results}
\label{subsec:main_results}
To demonstrate the effectiveness of AdaTKG, we present 
two complementary views of performance:

\textbf{(1) Comparison with various methods across emerging entities.}
Following prior work on inductive TKG~\cite{transfir,InGram}, we focus our comparison on emerging entities.
Table~\ref{tab:main_results} demonstrates that 
our method
outperforms all baselines across four benchmarks on link-prediction tasks.
This is consistent with our design, 
where AdaTKG learns \textit{how to integrate new interactions online} rather than storing 
learnable memory for seen entities, 
so even unseen entities accumulate useful signal at inference time.

\textbf{(2) Comparison with SoTA across emerging/unknown entities.}
Table~\ref{tab:sota_comparison} compares AdaTKG with TransFIR~\cite{transfir} across the two slices of unseen entities (\textit{Emerging}, \textit{Unknown}), jointly characterizing model behavior on cold-start and few-shot regimes. AdaTKG consistently outperforms TransFIR on both slices across all four benchmarks, indicating that the gain extends beyond the strict cold-start setting to entities 
merely seen rarely during training.

\begin{table*}[t]
\begin{minipage}{\linewidth}
\centering
\caption{\textbf{[Ablation 1] Paradigm grid.} The $2\!\times\!2$ grid of Static/Adaptive $\times$ Transductive/Inductive.}
\label{tab:ablation_paradigm}
\vspace{-3.5pt}
\adjustbox{max width=\linewidth}{
\footnotesize
\begin{tabular}{cc|ccc|ccc|ccc|ccc}
\toprule
\multicolumn{2}{c|}{\multirow{2.5}{*}{\textbf{Paradigm}}}
& \multicolumn{3}{c|}{\textbf{ICEWS14}}
& \multicolumn{3}{c|}{\textbf{ICEWS18}}
& \multicolumn{3}{c|}{\textbf{ICEWS05-15}}
& \multicolumn{3}{c}{\textbf{GDELT}} \\
\cmidrule(lr){3-5}\cmidrule(lr){6-8}\cmidrule(lr){9-11}\cmidrule(lr){12-14}
& 
& MRR & H@3 & H@10
& MRR & H@3 & H@10
& MRR & H@3 & H@10
& MRR & H@3 & H@10 \\
\midrule
\textcolor{secondcolor}{\textbf{Static}}   & \textcolor{secondcolor}{\textbf{Transductive}}      & .1478 & .1795 & .3147 & .0930 & .1016 & .2002 & .2084 & .2517 & .3663 
& .0866 & .0835 & .1883 \\
\textcolor{firstcolor}{\textbf{Adaptive}}  & \textcolor{secondcolor}{\textbf{Transductive}}      & .1612 & .1872 & .3236 & .1172 & .1320 & .2175 & .2129 & .2455 & .3652 & .0894 & .0940 & .1868 \\
\textcolor{secondcolor}{\textbf{Static}}   & \textcolor{firstcolor}{\textbf{Inductive}}          & .1763 & .2096 & .3413 & .1114 & .1230 & .2252 & .2177 & .2530 & .3708 & .1013 & .0994 & .2131 \\
\cellcolor{yellow!15}\textcolor{firstcolor}{\textbf{Adaptive}}
& \cellcolor{yellow!15}\textcolor{firstcolor}{\textbf{Inductive}}
& \cellcolor{yellow!15}\first{.2011}
& \cellcolor{yellow!15}\first{.2250}
& \cellcolor{yellow!15}\first{.3621}
& \cellcolor{yellow!15}\first{.1379}
& \cellcolor{yellow!15}\first{.1543}
& \cellcolor{yellow!15}\first{.2612}
& \cellcolor{yellow!15}\first{.2270}
& \cellcolor{yellow!15}\first{.2573}
& \cellcolor{yellow!15}\first{.3850}
& \cellcolor{yellow!15}\first{.1051}
& \cellcolor{yellow!15}\first{.1129}
& \cellcolor{yellow!15}\first{.2100} \\
\bottomrule
\end{tabular}
}
\end{minipage}

\vspace{13pt}
\begin{minipage}{\linewidth}
\centering
\caption{\textbf{[Ablation 2] Adaptive gate.} Effect of an adaptive gate $g_e$ vs. a constant gate $g=0.5$.}
\label{tab:ablation_gate}
\vspace{-3.5pt}
\adjustbox{max width=\linewidth}{
\footnotesize
\begin{tabular}{l|ccc|ccc|ccc|ccc}
\toprule
\multicolumn{1}{c|}{\multirow{2.5}{*}{\textbf{Gate}}}
& \multicolumn{3}{c|}{\textbf{ICEWS14}}
& \multicolumn{3}{c|}{\textbf{ICEWS18}}
& \multicolumn{3}{c|}{\textbf{ICEWS05-15}}
& \multicolumn{3}{c}{\textbf{GDELT}} \\
\cmidrule(lr){2-4}\cmidrule(lr){5-7}\cmidrule(lr){8-10}\cmidrule(lr){11-13}
& MRR & H@3 & H@10
& MRR & H@3 & H@10
& MRR & H@3 & H@10
& MRR & H@3 & H@10 \\
\midrule
w/o Adaptive gate
&        .1839  &        .2138  &        .3544  &        .1291  &        .1440  &        .2426  &        .2192  & .2530 &        .3850  &        .0982  &        .1032  &        .1976  \\
\cellcolor{yellow!15}w/ Adaptive gate
& \cellcolor{yellow!15}\first{.2011}
& \cellcolor{yellow!15}\first{.2250}
& \cellcolor{yellow!15}\first{.3621}
& \cellcolor{yellow!15}\first{.1379}
& \cellcolor{yellow!15}\first{.1543}
& \cellcolor{yellow!15}\first{.2612}
& \cellcolor{yellow!15}\first{.2270}
& \cellcolor{yellow!15}\first{.2573}
& \cellcolor{yellow!15}\first{.3850}
& \cellcolor{yellow!15}\first{.1051}
& \cellcolor{yellow!15}\first{.1129}
& \cellcolor{yellow!15}\first{.2301} \\
\midrule
\textit{$\Delta$ (\%)}
& \textbf{\up{+9.4}}  & \textbf{\up{+5.2}}  & \textbf{\up{+2.2}}
& \textbf{\up{+6.8}}  & \textbf{\up{+7.2}}  & \textbf{\up{+7.7}}
& \textbf{\up{+3.6}}  & \textbf{\up{+1.7}}     & \textbf{\up{+0.0}}
& \textbf{\up{+7.0}}  & \textbf{\up{+9.4}}  & \textbf{\up{+16.4}} \\
\bottomrule
\end{tabular}
}
\end{minipage}

\vspace{13pt}
\begin{minipage}{\linewidth}
\centering
\caption{\textbf{[Ablation 3] Update operator $\mathcal{U}$.} Three instantiations of the memory update rule.
}
\label{tab:ablation_operator}
\vspace{-3.5pt}
\adjustbox{max width=\linewidth}{
\footnotesize
\begin{tabular}{l|l|ccc|ccc|ccc|ccc}
\toprule
\multicolumn{2}{c}{\multirow{2.5}{*}{\textbf{Update operator}}}
& \multicolumn{3}{|c|}{\textbf{ICEWS14}}
& \multicolumn{3}{c|}{\textbf{ICEWS18}}
& \multicolumn{3}{c|}{\textbf{ICEWS05-15}}
& \multicolumn{3}{c}{\textbf{GDELT}} \\
\cmidrule(lr){3-5}\cmidrule(lr){6-8}\cmidrule(lr){9-11}\cmidrule(lr){12-14}
\multicolumn{2}{c|}{} & MRR & H@3 & H@10
& MRR & H@3 & H@10
& MRR & H@3 & H@10
& MRR & H@3 & H@10 \\
\midrule
\multicolumn{2}{c|}{Base (w/o Adaptivity)~\cite{transfir}} & .1763 & .2096 & .3413 & .1114 & .1230 & .2252 & .2177 & .2530 & .3708 & .1013 & .0994 & .2131 \\
\midrule
\multirow{13}{*}{AdaTKG}
& \cellcolor{yellow!15}EMA (\textit{default})
& \cellcolor{yellow!15}\first{.2011}
& \cellcolor{yellow!15}\first{.2250}
& \cellcolor{yellow!15}\first{.3621}
& \cellcolor{yellow!15}\first{.1379}
& \cellcolor{yellow!15}\first{.1543}
& \cellcolor{yellow!15}\first{.2612}
& \cellcolor{yellow!15}\first{.2270}
& \cellcolor{yellow!15}\first{.2573}
& \cellcolor{yellow!15}\first{.3850}
& \cellcolor{yellow!15}\first{.1051}
& \cellcolor{yellow!15}\first{.1129}
& \cellcolor{yellow!15}\first{.2301} \\
& \textit{$\Delta$ over Base (\%)}
& \textbf{\up{+14.1}} & \textbf{\up{+7.3}}  & \textbf{\up{+6.1}}
& \textbf{\up{+23.8}} & \textbf{\up{+25.4}} & \textbf{\up{+16.0}}
& \textbf{\up{+4.3}}  & \textbf{\up{+1.7}}  & \textbf{\up{+3.8}}
& \textbf{\up{+3.8}}  & \textbf{\up{+13.6}} & \textbf{\up{+8.0}}  \\
\cmidrule{2-14}
& \cellcolor{yellow!15}GRU
& \cellcolor{yellow!15}\first{.1955}
& \cellcolor{yellow!15}\first{.2192}
& \cellcolor{yellow!15}\first{.3582}
& \cellcolor{yellow!15}\first{.1428}
& \cellcolor{yellow!15}\first{.1599}
& \cellcolor{yellow!15}\first{.2605}
& \cellcolor{yellow!15}\first{.2330}
& \cellcolor{yellow!15}\first{.2700}
& \cellcolor{yellow!15}\first{.3925}
& \cellcolor{yellow!15}\first{.1112}
& \cellcolor{yellow!15}\first{.1141}
& \cellcolor{yellow!15}\first{.2243} \\
& \textit{$\Delta$ over Base (\%)}
& \textbf{\up{+10.9}} & \textbf{\up{+4.6}}  & \textbf{\up{+4.9}}
& \textbf{\up{+28.2}} & \textbf{\up{+30.0}} & \textbf{\up{+15.7}}
& \textbf{\up{+7.0}}  & \textbf{\up{+6.7}}  & \textbf{\up{+5.9}}
& \textbf{\up{+9.8}}  & \textbf{\up{+14.8}} & \textbf{\up{+5.3}}  \\
\cmidrule{2-14}
& \cellcolor{yellow!15}Cross-attention
& \cellcolor{yellow!15}\first{.1913}
& \cellcolor{yellow!15}\first{.2296}
& \cellcolor{yellow!15}\first{.3609}
& \cellcolor{yellow!15}\first{.1454}
& \cellcolor{yellow!15}\first{.1712}
& \cellcolor{yellow!15}\first{.2761}
& \cellcolor{yellow!15}\first{.2243}
& \cellcolor{yellow!15}\first{.2573}
& \cellcolor{yellow!15}\first{.3815}
& \cellcolor{yellow!15}\first{.1297}
& \cellcolor{yellow!15}\first{.1396}
& \cellcolor{yellow!15}\first{.2544} \\
& \textit{$\Delta$ over Base (\%)}
& \textbf{\up{+8.5}}  & \textbf{\up{+9.5}}  & \textbf{\up{+5.7}}
& \textbf{\up{+30.5}} & \textbf{\up{+39.2}} & \textbf{\up{+22.6}}
& \textbf{\up{+3.0}}  & \textbf{\up{+1.7}}  & \textbf{\up{+2.9}}
& \textbf{\up{+28.0}} & \textbf{\up{+40.4}} & \textbf{\up{+19.4}} \\
\bottomrule
\end{tabular}
}
\end{minipage}

\vspace{13pt}
\begin{minipage}{\linewidth}
\centering
\caption{\textbf{[Ablation 4] EMA decay parameter.} Three parameterizations of the EMA decay rate.}
\label{tab:ablation_ema}
\vspace{-3.5pt}
\adjustbox{max width=\linewidth}{
\footnotesize
\begin{tabular}{l|l|ccc|ccc|ccc|ccc}
\toprule
\multicolumn{2}{c}{\multirow{2.5}{*}{\textbf{EMA decay rate}}}
& \multicolumn{3}{|c|}{\textbf{ICEWS14}}
& \multicolumn{3}{c|}{\textbf{ICEWS18}}
& \multicolumn{3}{c|}{\textbf{ICEWS05-15}}
& \multicolumn{3}{c}{\textbf{GDELT}} \\
\cmidrule(lr){3-5}\cmidrule(lr){6-8}\cmidrule(lr){9-11}\cmidrule(lr){12-14}
\multicolumn{2}{c|}{} & MRR & H@3 & H@10
& MRR & H@3 & H@10
& MRR & H@3 & H@10
& MRR & H@3 & H@10 \\
\midrule
\multicolumn{2}{c|}{Base (w/o Adaptivity)~\cite{transfir}} & .1763 & .2096 & .3413 & .1114 & .1230 & .2252 & .2177 & .2530 & .3708 & .1013 & .0994 & .2131 \\
\midrule
Shared scalar & $\alpha\!=\!\sigma(\rho)$            & \first{.2011} & \first{.2250} & \first{.3621} & .1379         & \first{.1543} & .2612         & \first{.2270} & \first{.2573} & \first{.3850} & \first{.1051} & \first{.1129} & \first{.2301} \\
Per-entity scalar & $\alpha_e\!=\!\sigma(\rho_e)$    & .1888         & .2165         & .3525         & \first{.1389} & .1529         & \first{.2615} & .2199         & .2562         & .3808         & .1008         & .1056         & .2046         \\
Per-dim. vector & $\boldsymbol{\alpha}\!=\!\sigma(\boldsymbol{\rho})$ & .1885 & .2192 & .3532 & .1358 & .1462 & .2556 & .2129 & .2442 & .3753 & .0990 & .1052 & .2057 \\
\bottomrule
\end{tabular}
}
\end{minipage}
\end{table*}

\subsection{Ablation Study}
\label{subsec:ablation}
To isolate the contribution of each design choice in AdaTKG, we perform a controlled ablation on the emerging slice across all four benchmarks, organized along four axes:

\textbf{(1) Effectiveness of Adaptive + Inductive paradigm.}
The four rows in Table~\ref{tab:ablation_paradigm} traverse the $2\!\times\!2$ paradigm grid: \textit{Static + Inductive} ($\equiv$ TransFIR~\cite{transfir}) keeps only the cluster prior, \textit{Adaptive + Transductive} keeps only the per-entity memory, and \textit{Adaptive + Inductive} is the AdaTKG that combines both.
Removing the per-entity memory collapses AdaTKG to TransFIR, 
while removing the cluster prior breaks the cold-start match with TransFIR.
The proposed Adaptive + Inductive paradigm achieves the best results, demonstrating the effectiveness of adaptivity in TKG reasoning.

\textbf{(2) Effectiveness of adaptive gate.}
As shown in Table~\ref{tab:ablation_gate}, replacing the learned gate by the constant $g = 0.5$ removes 
its ability to track the varying reliability of the memory and degrades performance, 
confirming that the gate is a necessary mechanism for the graceful cold-start behavior.

\textbf{(3) Robustness to update operator.}
We compare three instantiations of the stateful update operator $\mathcal{U}$,
the 1) learnable EMA, a 2) GRU cell, and a 
3) cross-attention readout over a bounded per-entity buffer.
All three retain the adaptive gate of Eq.~\eqref{eq:gate} for fusion, so \textit{only the memory update rule itself differs}.
As shown in Table~\ref{tab:ablation_operator}, every variant improves over the static baseline~\cite{transfir}, 
indicating that the gain is driven by the \textit{adaptive property} of per-entity memory rather than by any particular implementation.
It is important to note that our focus is \textit{not on the choice of EMA itself but on this adaptive property}, 
and we adopt EMA as a lightweight yet effective realization of $\mathcal{U}$.

\textbf{(4) Variants of EMA decay parameterization.}
The default AdaTKG uses a single shared learnable scalar $\alpha\!=\!\sigma(\rho)$ for the EMA decay.
Two natural relaxations enlarge the parameterization: 
a 1)~\textit{per-entity scalar}\footnote{This breaks inductivity for emerging entities, where we fall back to the cluster-prototype decay $\rho_{\pi(e)}$.} $\alpha_e\!=\!\sigma(\rho_e)$ that lets each entity learn its own forgetting rate
and a 2)~\textit{per-dimension vector} $\boldsymbol{\alpha}\!=\!\sigma(\boldsymbol{\rho})\!\in\![0,1]^d$ 
that decays each coordinate at a different rate.
As shown in Table~\ref{tab:ablation_ema}, while 
all designs 
outperform the baseline (w/o Adaptivity), the shared scalar 
outperforms both relaxations, indicating that a \textit{single shared scalar} already suffices for the memory branch.

\begin{figure}[t]
\centering
\begin{adjustbox}{max width=\linewidth}
\includegraphics[width=\linewidth]{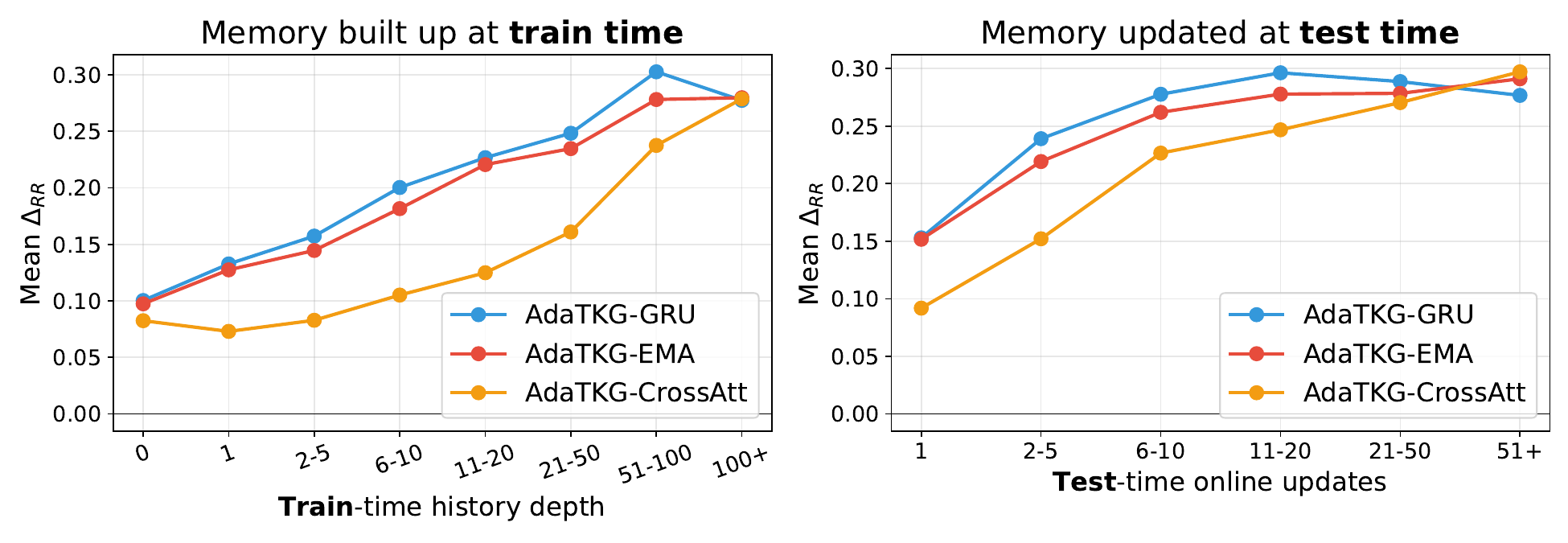}
\end{adjustbox}
\caption{\textbf{Performance by \# interactions at train and test time.}
We measure the memory's contribution by $\Delta_{RR} = \mathrm{RR}_{\mathrm{full}} - \mathrm{RR}_{\mathrm{zero}}$ (i.e., the difference between the per-query RR of the model 
w/ and w/o
its memory branch) and stratify it along two axes: \textbf{(Left)} \textbf{Train}-time history depth of the subject and the \textbf{(Right)} \textbf{Test}-time online updates accumulated during inference. The increase of $\Delta_{RR}$ along both axes, consistent across all three operators, indicates that more observed interactions translate 
into a \textit{larger memory contribution} and, in turn, into \textit{higher predictive performance}.
}
\label{fig:memory_ablation}
\vspace{-9pt}
\end{figure}

\begin{figure}[t]
\centering
\begin{adjustbox}{max width=\linewidth}
\includegraphics[width=\linewidth]{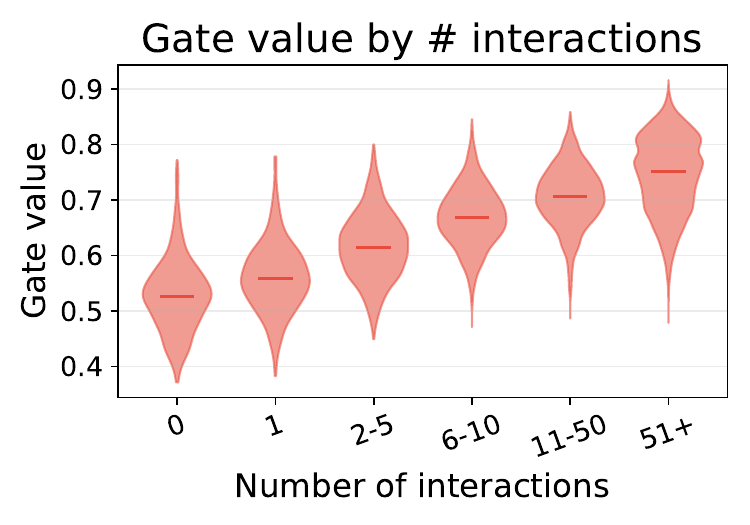}
\end{adjustbox}
\caption{\textbf{Gate value by \# interactions.} Distribution of the learned gate $g^{(t_q)}_e$,
stratified by the number of interactions.}
\label{fig:gate}
\end{figure}

\subsection{Analysis}
\label{subsec:analysis}
\textbf{(1) Effectiveness of memory by \# interactions. (Figure~\ref{fig:memory_ablation}).}
To demonstrate that the \textit{per-entity memory is the actual source of AdaTKG's gain}, we score each AdaTKG variant on ICEWS14 in two configurations:
1) with the trained memory branch active ($\mathrm{RR}_{\mathrm{full}}$), and 
2) with the memory branch bypassed by setting the gate to zero ($\mathrm{RR}_{\mathrm{zero}}$).
The per-query difference $\Delta_{RR} = \mathrm{RR}_{\mathrm{full}} - \mathrm{RR}_{\mathrm{zero}}$ is thus \textit{the performance gain by the memory branch}, and we stratify it along two complementary axes in Figure~\ref{fig:memory_ablation}: (Left) train-time history depth of the subject and (Right) test-time online updates accumulated during inference.
$\Delta_{RR}$ is positive across every bin and grows along both axes for all three update operators, 
confirming that more interactions translate into a larger memory contribution and a corresponding gain in MRR.
Results across all four benchmarks are provided in Appendix~\ref{app:memory_ablation_others}.

\textbf{(2) Gate activation by \# interactions (Figure~\ref{fig:gate}).}
The adaptive gate $g^{(t_q)}_e$ controls how much of the representation comes from the \textit{static inductive prior} versus the \textit{per-entity memory}, and its trajectory offers a direct window into what the model has learned.
As shown in Figure~\ref{fig:gate} with ICEWS14 and the EMA update operator, 
the gate value grows
with the number of observed interactions.
See Appendix~\ref{app:gate_violin_grid} for results across all benchmarks and update operators.

\begin{figure}[t]
\centering
\begin{adjustbox}{max width=\linewidth}
\includegraphics[width=\linewidth]{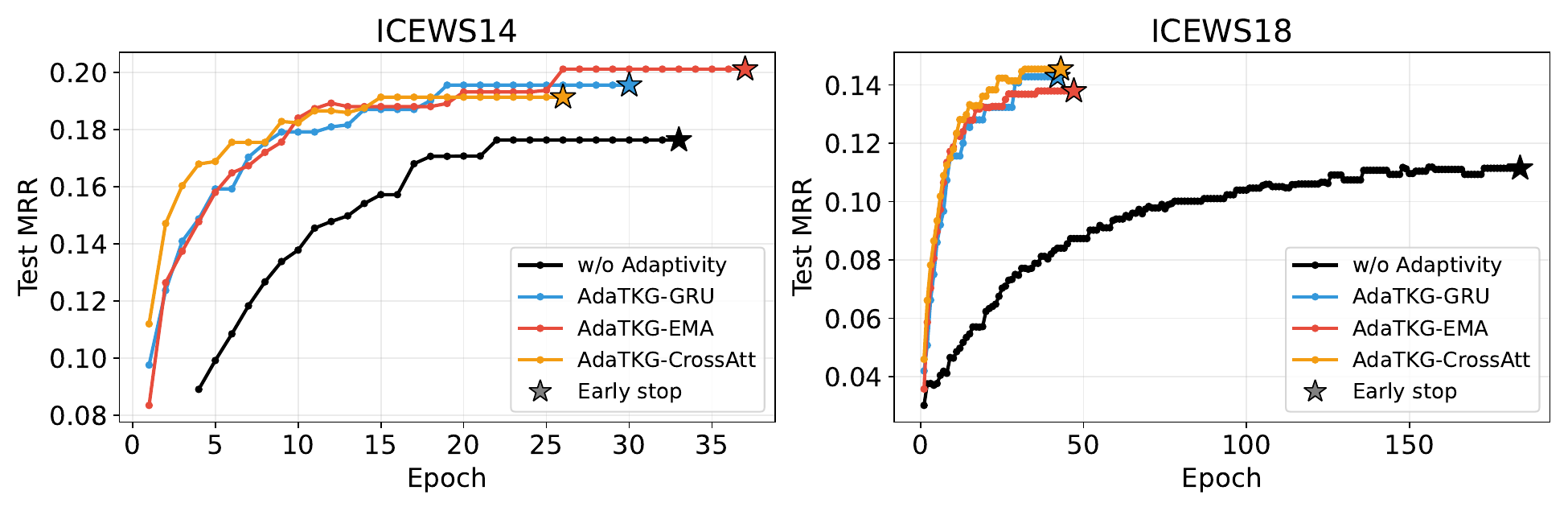}
\end{adjustbox}
\caption{\textbf{Training curves on test MRR.}
Per-epoch test emerging MRR for the static-inductive baseline (\textit{w/o Adaptivity}) and the three AdaTKG update operators (\textit{w/ Adaptivity}).
AdaTKG update operators yield faster convergence and higher performance.}
\label{fig:training_curves}
\vspace{-5pt}
\end{figure}

\textbf{(3) Faster convergence and higher final performance (Figure~\ref{fig:training_curves}).}
Figure~\ref{fig:training_curves} traces the per-epoch test emerging MRR
for the static-inductive baseline (\textit{w/o Adaptivity}) and the three AdaTKG update operators.
All three 
variants 1) \textit{converge in fewer epochs} and to a 2) \textit{higher final MRR} than the baseline.
This indicates that adaptivity yields both 1)~\textit{training efficiency} and 2)~\textit{predictive performance}, and that these gains stem from the 
memory itself rather than from any specific update operator.

\begin{table}[t]
\centering
\caption{\textbf{Robustness to training horizon.} \textit{Training horizon} ($k\%$) denotes the fraction of the most recent training time window used; AdaTKG's gain over the baseline persists at every horizon and is largest at the shortest.}
\label{tab:horizon}
\begin{adjustbox}{max width=\linewidth}
\small
\begin{tabular}{l|cccc}
\toprule
 & \multicolumn{4}{c}{\textbf{Training horizon}}\\
 \cmidrule(lr){2-5}
 & 25\% & 50\% & 75\% & 100\% \\
\midrule
Base (w/o Adaptivity)~\cite{transfir}
& .1089 & .1408 & .1357 & .1763 \\
\cellcolor{yellow!15}\textbf{AdaTKG}
& \cellcolor{yellow!15}\first{.1435}
& \cellcolor{yellow!15}\first{.1676}
& \cellcolor{yellow!15}\first{.1727}
& \cellcolor{yellow!15}\first{.2011} \\
\midrule
\textit{$\Delta$ over SoTA (\%)}     & \textbf{\up{+31.8}} & \textbf{\up{+19.0}} & \textbf{\up{+27.3}} & \textbf{\up{+14.1}} \\
\bottomrule
\end{tabular}
\end{adjustbox}
\vspace{8pt}
\end{table}

\begin{table*}[t]
\centering
\caption{\textbf{Efficiency comparison.}
We compare TransFIR and the three AdaTKG update operators along 1) the number of parameters, 2) the training time per epoch, and 3) the FLOPs per query.
AdaTKG delivers a substantial performance gain over TransFIR with only marginal extra computation.}
\label{tab:efficiency}
\adjustbox{max width=\linewidth}{
\small
\begin{tabular}{c|l|cc|cc|cc|ccc}
\toprule
\multicolumn{2}{c|}{\multirow{4}{*}{\textbf{Method}}}
  & \multicolumn{6}{c|}{\textbf{[1] Efficiency}}
  & \multicolumn{3}{c}{\textbf{[2] Performance}} \\
\cmidrule(lr){3-8}\cmidrule(lr){9-11}
\multicolumn{2}{c|}{}
  & \multicolumn{2}{c|}{\textbf{\# Parameters} (M)}
  & \multicolumn{2}{c|}{\textbf{Training time} (s/epoch)}
  & \multicolumn{2}{c|}{\textbf{FLOPs} (M/query)}
  & \multirow{2.5}{*}{\textbf{MRR}} & \multirow{2.5}{*}{\textbf{H@3}} & \multirow{2.5}{*}{\textbf{H@10}} \\
\cmidrule(lr){3-4}\cmidrule(lr){5-6}\cmidrule(lr){7-8}
\multicolumn{2}{c|}{}
  & Value & $\Delta$\,(\%)
  & Value & $\Delta$\,(\%)
  & Value & $\Delta$\,(\%)
  &       &       & \\
\midrule
\multicolumn{2}{l|}{Base (w/o Adaptivity)~\cite{transfir}}
  & 41.15  & --
  & 56.1   & --
  & 399.8  & --
  & .1763  & .2096 & .3413 \\
\midrule
\multirow{3}{*}{\textbf{AdaTKG}}
  & EMA (\textit{default})
  & 44.10  & \textbf{+7.2\%}
  & 64.1   & \textbf{+14.3\%}
  & 405.7  & \textbf{+1.5\%}
  & \first{.2011} & \first{.2250} & \first{.3621} \\
  & GRU
  & 47.65  & \textbf{+15.8\%}
  & 67.6   & \textbf{+20.5\%}
  & 408.8  & \textbf{+2.2\%}
  & \first{.1955} & \first{.2192} & \first{.3582} \\
  & Cross-attention
  & 46.47  & \textbf{+12.9\%}
  & 62.7   & \textbf{+11.8\%}
  & 445.9  & \textbf{+11.5\%}
  & \first{.1913} & \first{.2296} & \first{.3609} \\
\bottomrule
\end{tabular}
}
\end{table*}

\textbf{(4) Robustness to training horizon (Table~\ref{tab:horizon}).}
We additionally train both methods on only the last $k\%$ of the training time window 
to test whether the gain depends on long training history. AdaTKG outperforms TransFIR at every horizon, 
with the 
gain
\textit{largest at the shortest horizon}. This is consistent with the asymmetric roles of the two branches: the static branch is \textit{parametric} and its quality scales with how much training data it sees, while the memory branch is \textit{non-parametric} and accumulates online from the test stream, so its marginal value is highest when the static prior is weakest.

\begin{figure}[t]
\centering
\begin{adjustbox}{max width=\linewidth}
\includegraphics[width=\linewidth]{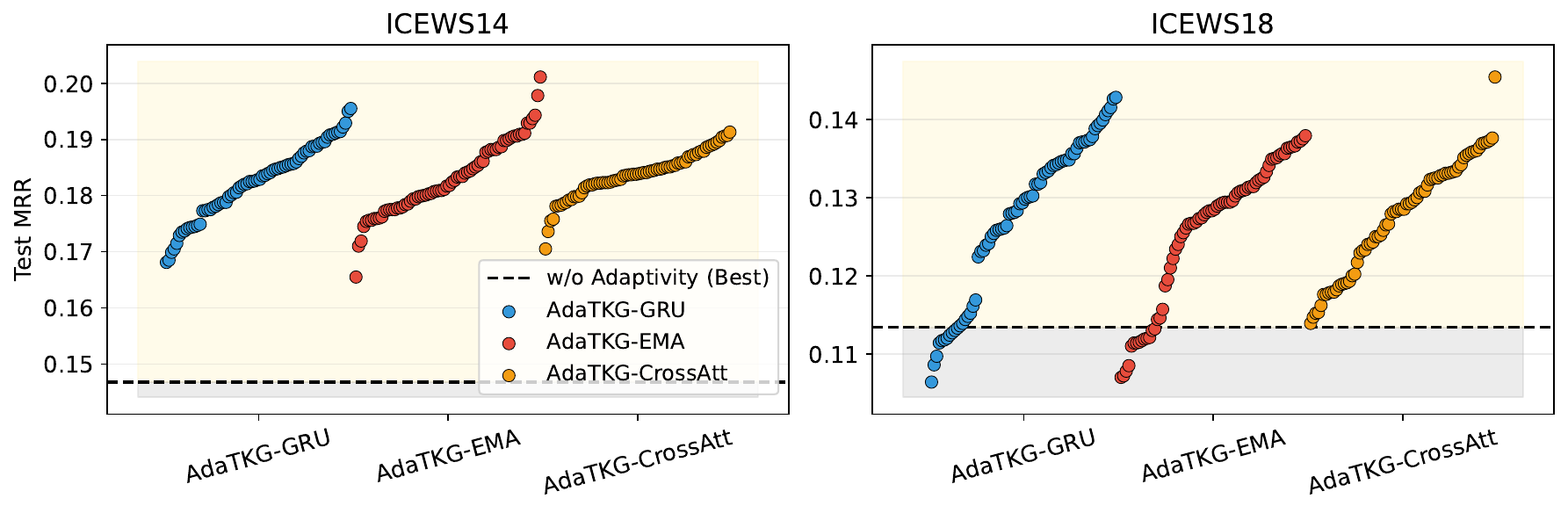}
\end{adjustbox}
\caption{\textbf{Sensitivity to memory-related hyperparameters.} Each point is one hyperparameter setting of AdaTKG, compared against the \textit{w/o Adaptivity} baseline at its best setting (dashed line).}
\label{fig:hp_sensitivity}
\end{figure}

\textbf{(5) Sensitivity to memory-related hyperparameters (Figure~\ref{fig:hp_sensitivity}).}
For each of ICEWS14 and ICEWS18, we sweep the full hyperparameter grid (chain length, hidden dim, \# layers, codebook size) and plot the resulting test emerging MRR for every AdaTKG operator 
against the best static-inductive baseline. Every AdaTKG HP point lies above the baseline reference line on both benchmarks, indicating that the adaptive memory mechanism is \textit{robust to hyperparameter choices}.

\textbf{(6) Efficiency analysis (Table~\ref{tab:efficiency}).}
We compare AdaTKG with TransFIR,
the SoTA static-inductive baseline,
on three computational dimensions on ICEWS14:
1) number of parameters, 2) training time per epoch, and 3) FLOPs per query,
evaluating all four methods under the same hyperparameter configuration 
(= AdaTKG-EMA's best HP on ICEWS14) 
for fair comparison.
As shown in Table~\ref{tab:efficiency}, the three update operators all add only a \textit{small overhead} over the base while \textit{substantially improving performance}, confirming that adaptivity is the source of the gain rather than added model capacity.
Among them, we adopt EMA as our default operator as it offers the best balance between efficiency and performance.
Comparison on the other three benchmarks is provided in Appendix~\ref{app:efficiency_others}.

\begin{figure}[t]
\centering
\begin{adjustbox}{max width=\linewidth}
\includegraphics[width=\linewidth]{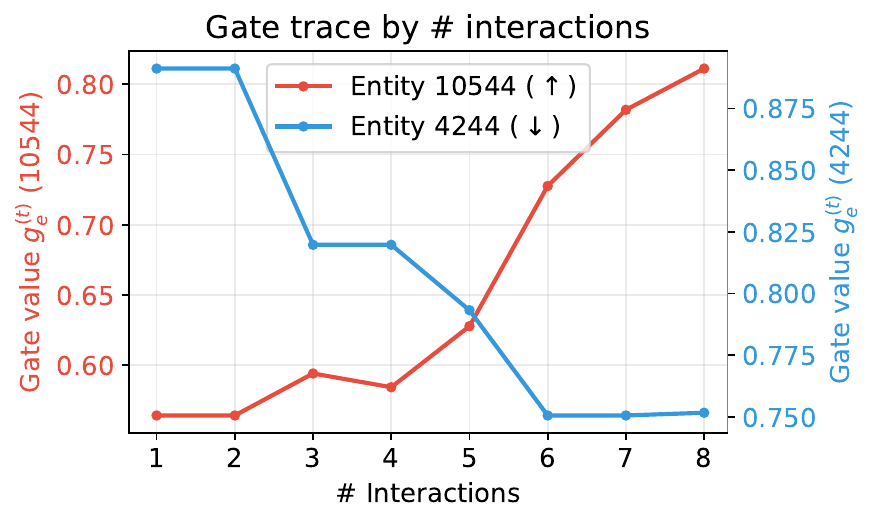}
\end{adjustbox}
\caption{
Qualitative gate trace.
Test-time gate $g_e^{(t)}$ of two entities: one rises as it departs from the static prior (memory-reliant, red), the other stays low (prior-reliant, blue)
}
\label{fig:qualitative}
\end{figure}

\textbf{(7) Qualitative ex: How memory adapts (Figure~\ref{fig:qualitative}).}
While Figure~\ref{fig:gate} reports the \textit{average} gate trajectory, here we trace the gate $g_e^{(t)}$ of two individual entities on ICEWS18 with AdaTKG across their successive test-time appearances. The red trajectory ($\uparrow$) corresponds to an entity whose gate climbs rapidly, indicating that its observed actions diverge from the \textit{static inductive prior} and the model accordingly leans on the \textit{per-entity memory branch}; the blue trajectory ($\downarrow$) corresponds to an entity whose gate remains low throughout, indicating that the \textit{static inductive prior} already explains its behavior well and the model accordingly trusts it. This entity-level contrast confirms that the gate is not a function of time or interaction count alone, but of the \textit{entity-specific deviation} from the \textit{static prior}, which is the behavior anticipated by the design.

\section{Conclusion}
\label{sec:conclusion}

We propose AdaTKG, which shifts TKG reasoning 
from \textit{static} to \textit{adaptive} representations 
by equipping every entity with a memory governed by an adaptive gate.
Crucially, the memory is treated as an internal state rather than a learnable parameter, 
with all learnable capacity concentrated in a single shared update rule, 
making it
applicable to unseen entities.
Across various datasets, AdaTKG yields consistent improvements, with the gain growing in the number of interactions per entity.

\textbf{Limitation and Future Work.}
AdaTKG's per-entity memory must be persisted between inference calls, introducing a simple bookkeeping requirement absent in stateless inductive methods.
Coupling the memory with a continually adapting codebook lets the static prior itself accommodate slow distribution drift on long-horizon TKGs, which we view as a natural next step.

\FloatBarrier   


\section*{Impact Statement}
AdaTKG targets a practically consequential setting: reasoning about entities that continuously emerge in temporal knowledge graphs without any training-time footprint, which can broaden analyses in domains such as financial monitoring and public-health surveillance.
Because event-based TKGs inherit reporting biases that may propagate into AdaTKG's per-entity memories, its outputs should serve as decision support validated by human experts.

\section*{GenAI Usage Disclosure}
Generative AI tools were used in preparing this paper solely for light editing, such as grammar checking and minor wording refinements, and not to generate research ideas, design or conduct experiments, produce results, or write any substantive technical content, all of which was authored and verified by the authors.


\bibliographystyle{ACM-Reference-Format}
\bibliography{references}


\begin{thebibliography}{42}


\ifx \showCODEN    \undefined \def \showCODEN     #1{\unskip}     \fi
\ifx \showISBNx    \undefined \def \showISBNx     #1{\unskip}     \fi
\ifx \showISBNxiii \undefined \def \showISBNxiii  #1{\unskip}     \fi
\ifx \showISSN     \undefined \def \showISSN      #1{\unskip}     \fi
\ifx \showLCCN     \undefined \def \showLCCN      #1{\unskip}     \fi
\ifx \shownote     \undefined \def \shownote      #1{#1}          \fi
\ifx \showarticletitle \undefined \def \showarticletitle #1{#1}   \fi
\ifx \showURL      \undefined \def \showURL       {\relax}        \fi
\providecommand\bibfield[2]{#2}
\providecommand\bibinfo[2]{#2}
\providecommand\natexlab[1]{#1}
\providecommand\showeprint[2][]{arXiv:#2}

\bibitem[Boschee et~al\mbox{.}(2015)]%
        {DVN_28075_2015}
\bibfield{author}{\bibinfo{person}{Elizabeth Boschee}, \bibinfo{person}{Jennifer Lautenschlager}, \bibinfo{person}{Sean O'Brien}, \bibinfo{person}{Steve Shellman}, \bibinfo{person}{James Starz}, {and} \bibinfo{person}{Michael Ward}.} \bibinfo{year}{2015}\natexlab{}.
\newblock \bibinfo{title}{{ICEWS Coded Event Data}}.
\newblock
\href{https://doi.org/10.7910/DVN/28075}{doi:\nolinkurl{10.7910/DVN/28075}}


\bibitem[Cai et~al\mbox{.}(2023)]%
        {tkgsurvey_1}
\bibfield{author}{\bibinfo{person}{Borui Cai}, \bibinfo{person}{Yong Xiang}, \bibinfo{person}{Longxiang Gao}, \bibinfo{person}{He Zhang}, \bibinfo{person}{Yunfeng Li}, {and} \bibinfo{person}{Jianxin Li}.} \bibinfo{year}{2023}\natexlab{}.
\newblock \showarticletitle{Temporal knowledge graph completion: a survey}. In \bibinfo{booktitle}{\emph{Proceedings of the Thirty-Second International Joint Conference on Artificial Intelligence}}. \bibinfo{pages}{6545--6553}.
\newblock


\bibitem[Chen et~al\mbox{.}(2021)]%
        {TACT}
\bibfield{author}{\bibinfo{person}{Jiajun Chen}, \bibinfo{person}{Huarui He}, \bibinfo{person}{Feng Wu}, {and} \bibinfo{person}{Jie Wang}.} \bibinfo{year}{2021}\natexlab{}.
\newblock \showarticletitle{Topology-aware correlations between relations for inductive link prediction in knowledge graphs}. In \bibinfo{booktitle}{\emph{Proceedings of the AAAI conference on artificial intelligence}}, Vol.~\bibinfo{volume}{35}. \bibinfo{pages}{6271--6278}.
\newblock


\bibitem[Chen et~al\mbox{.}(2022)]%
        {morse}
\bibfield{author}{\bibinfo{person}{Mingyang Chen}, \bibinfo{person}{Wen Zhang}, \bibinfo{person}{Yushan Zhu}, \bibinfo{person}{Hongting Zhou}, \bibinfo{person}{Zonggang Yuan}, \bibinfo{person}{Changliang Xu}, {and} \bibinfo{person}{Huajun Chen}.} \bibinfo{year}{2022}\natexlab{}.
\newblock \showarticletitle{Meta-knowledge transfer for inductive knowledge graph embedding}. In \bibinfo{booktitle}{\emph{Proceedings of the 45th international ACM SIGIR conference on research and development in information retrieval}}. \bibinfo{pages}{927--937}.
\newblock


\bibitem[Chen et~al\mbox{.}(2024)]%
        {LogCL}
\bibfield{author}{\bibinfo{person}{Wei Chen}, \bibinfo{person}{Huaiyu Wan}, \bibinfo{person}{Yuting Wu}, \bibinfo{person}{Shuyuan Zhao}, \bibinfo{person}{Jiayaqi Cheng}, \bibinfo{person}{Yuxin Li}, {and} \bibinfo{person}{Youfang Lin}.} \bibinfo{year}{2024}\natexlab{}.
\newblock \showarticletitle{Local-global history-aware contrastive learning for temporal knowledge graph reasoning}. In \bibinfo{booktitle}{\emph{2024 IEEE 40th International Conference on Data Engineering (ICDE)}}. IEEE, \bibinfo{pages}{733--746}.
\newblock


\bibitem[Cho et~al\mbox{.}(2014)]%
        {cho2014gru}
\bibfield{author}{\bibinfo{person}{Kyunghyun Cho}, \bibinfo{person}{Bart van Merri{\"e}nboer}, \bibinfo{person}{Caglar Gulcehre}, \bibinfo{person}{Dzmitry Bahdanau}, \bibinfo{person}{Fethi Bougares}, \bibinfo{person}{Holger Schwenk}, {and} \bibinfo{person}{Yoshua Bengio}.} \bibinfo{year}{2014}\natexlab{}.
\newblock \showarticletitle{Learning Phrase Representations using {RNN} Encoder--Decoder for Statistical Machine Translation}. In \bibinfo{booktitle}{\emph{Proceedings of the 2014 Conference on Empirical Methods in Natural Language Processing (EMNLP)}}. \bibinfo{pages}{1724--1734}.
\newblock


\bibitem[Devlin et~al\mbox{.}(2019)]%
        {bert}
\bibfield{author}{\bibinfo{person}{Jacob Devlin}, \bibinfo{person}{Ming-Wei Chang}, \bibinfo{person}{Kenton Lee}, {and} \bibinfo{person}{Kristina Toutanova}.} \bibinfo{year}{2019}\natexlab{}.
\newblock \showarticletitle{Bert: Pre-training of deep bidirectional transformers for language understanding}. In \bibinfo{booktitle}{\emph{Proceedings of the 2019 conference of the North American chapter of the association for computational linguistics: human language technologies, volume 1 (long and short papers)}}. \bibinfo{pages}{4171--4186}.
\newblock


\bibitem[Ding et~al\mbox{.}(2024)]%
        {zrllm}
\bibfield{author}{\bibinfo{person}{Zifeng Ding}, \bibinfo{person}{Heling Cai}, \bibinfo{person}{Jingpei Wu}, \bibinfo{person}{Yunpu Ma}, \bibinfo{person}{Ruotong Liao}, \bibinfo{person}{Bo Xiong}, {and} \bibinfo{person}{Volker Tresp}.} \bibinfo{year}{2024}\natexlab{}.
\newblock \showarticletitle{zrllm: Zero-shot relational learning on temporal knowledge graphs with large language models}. In \bibinfo{booktitle}{\emph{Proceedings of the 2024 conference of the North American chapter of the association for computational linguistics: Human language technologies (Volume 1: Long papers)}}. \bibinfo{pages}{1877--1895}.
\newblock


\bibitem[Fang et~al\mbox{.}(2024)]%
        {ECEformer}
\bibfield{author}{\bibinfo{person}{Zhiyu Fang}, \bibinfo{person}{Shuai-Long Lei}, \bibinfo{person}{Xiaobin Zhu}, \bibinfo{person}{Chun Yang}, \bibinfo{person}{Shi-Xue Zhang}, \bibinfo{person}{Xu-Cheng Yin}, {and} \bibinfo{person}{Jingyan Qin}.} \bibinfo{year}{2024}\natexlab{}.
\newblock \showarticletitle{Transformer-based reasoning for learning evolutionary chain of events on temporal knowledge graph}. In \bibinfo{booktitle}{\emph{Proceedings of the 47th international ACM SIGIR conference on research and development in information retrieval}}. \bibinfo{pages}{70--79}.
\newblock


\bibitem[Galkin et~al\mbox{.}(2024)]%
        {ultra}
\bibfield{author}{\bibinfo{person}{Mikhail Galkin}, \bibinfo{person}{Xinyu Yuan}, \bibinfo{person}{Hesham Mostafa}, \bibinfo{person}{Jian Tang}, {and} \bibinfo{person}{Zhaocheng Zhu}.} \bibinfo{year}{2024}\natexlab{}.
\newblock \showarticletitle{Towards Foundation Models for Knowledge Graph Reasoning}. In \bibinfo{booktitle}{\emph{The Twelfth International Conference on Learning Representations}}.
\newblock


\bibitem[Garcia-Duran et~al\mbox{.}(2018)]%
        {TTransE}
\bibfield{author}{\bibinfo{person}{Alberto Garcia-Duran}, \bibinfo{person}{Sebastijan Duman{\v{c}}i{\'c}}, {and} \bibinfo{person}{Mathias Niepert}.} \bibinfo{year}{2018}\natexlab{}.
\newblock \showarticletitle{Learning Sequence Encoders for Temporal Knowledge Graph Completion}. In \bibinfo{booktitle}{\emph{Proceedings of the 2018 Conference on Empirical Methods in Natural Language Processing}}. \bibinfo{pages}{4816--4821}.
\newblock


\bibitem[Kingma and Ba(2015)]%
        {kingma2014adam}
\bibfield{author}{\bibinfo{person}{Diederik~P. Kingma} {and} \bibinfo{person}{Jimmy Ba}.} \bibinfo{year}{2015}\natexlab{}.
\newblock \showarticletitle{Adam: A Method for Stochastic Optimization}. In \bibinfo{booktitle}{\emph{International Conference on Learning Representations (ICLR)}}.
\newblock


\bibitem[Lacroix et~al\mbox{.}(2020)]%
        {TNTComplex}
\bibfield{author}{\bibinfo{person}{Timoth{\'e}e Lacroix}, \bibinfo{person}{Guillaume Obozinski}, {and} \bibinfo{person}{Nicolas Usunier}.} \bibinfo{year}{2020}\natexlab{}.
\newblock \showarticletitle{Tensor Decompositions for Temporal Knowledge Base Completion}. In \bibinfo{booktitle}{\emph{International Conference on Learning Representations}}.
\newblock


\bibitem[Lee et~al\mbox{.}(2023a)]%
        {ICL}
\bibfield{author}{\bibinfo{person}{Dong-Ho Lee}, \bibinfo{person}{Kian Ahrabian}, \bibinfo{person}{Woojeong Jin}, \bibinfo{person}{Fred Morstatter}, {and} \bibinfo{person}{Jay Pujara}.} \bibinfo{year}{2023}\natexlab{a}.
\newblock \showarticletitle{Temporal Knowledge Graph Forecasting Without Knowledge Using In-Context Learning}. In \bibinfo{booktitle}{\emph{Proceedings of the 2023 Conference on Empirical Methods in Natural Language Processing}}. \bibinfo{pages}{544--557}.
\newblock


\bibitem[Lee et~al\mbox{.}(2023b)]%
        {InGram}
\bibfield{author}{\bibinfo{person}{Jaejun Lee}, \bibinfo{person}{Chanyoung Chung}, {and} \bibinfo{person}{Joyce~Jiyoung Whang}.} \bibinfo{year}{2023}\natexlab{b}.
\newblock \showarticletitle{InGram: Inductive knowledge graph embedding via relation graphs}. In \bibinfo{booktitle}{\emph{International conference on machine learning}}. PMLR, \bibinfo{pages}{18796--18809}.
\newblock


\bibitem[Leetaru and Schrodt(2013)]%
        {leetaru2013gdelt}
\bibfield{author}{\bibinfo{person}{Kalev Leetaru} {and} \bibinfo{person}{Philip~A Schrodt}.} \bibinfo{year}{2013}\natexlab{}.
\newblock \showarticletitle{{GDELT}: Global Data on Events, Location, and Tone, 1979--2012}. In \bibinfo{booktitle}{\emph{ISA annual convention}}, Vol.~\bibinfo{volume}{2}. Citeseer, \bibinfo{pages}{1--49}.
\newblock


\bibitem[Li et~al\mbox{.}(2022)]%
        {Hismatch}
\bibfield{author}{\bibinfo{person}{Zixuan Li}, \bibinfo{person}{Zhongni Hou}, \bibinfo{person}{Saiping Guan}, \bibinfo{person}{Xiaolong Jin}, \bibinfo{person}{Weihua Peng}, \bibinfo{person}{Long Bai}, \bibinfo{person}{Yajuan Lyu}, \bibinfo{person}{Wei Li}, \bibinfo{person}{Jiafeng Guo}, {and} \bibinfo{person}{Xueqi Cheng}.} \bibinfo{year}{2022}\natexlab{}.
\newblock \showarticletitle{HiSMatch: Historical Structure Matching based Temporal Knowledge Graph Reasoning}. In \bibinfo{booktitle}{\emph{Findings of the Association for Computational Linguistics: EMNLP 2022}}. \bibinfo{pages}{7328--7338}.
\newblock


\bibitem[Li et~al\mbox{.}(2021)]%
        {REGCN}
\bibfield{author}{\bibinfo{person}{Zixuan Li}, \bibinfo{person}{Xiaolong Jin}, \bibinfo{person}{Wei Li}, \bibinfo{person}{Saiping Guan}, \bibinfo{person}{Jiafeng Guo}, \bibinfo{person}{Huawei Shen}, \bibinfo{person}{Yuanzhuo Wang}, {and} \bibinfo{person}{Xueqi Cheng}.} \bibinfo{year}{2021}\natexlab{}.
\newblock \showarticletitle{Temporal knowledge graph reasoning based on evolutional representation learning}. In \bibinfo{booktitle}{\emph{Proceedings of the 44th international ACM SIGIR conference on research and development in information retrieval}}. \bibinfo{pages}{408--417}.
\newblock


\bibitem[Liao et~al\mbox{.}(2024)]%
        {GenTKG}
\bibfield{author}{\bibinfo{person}{Ruotong Liao}, \bibinfo{person}{Xu Jia}, \bibinfo{person}{Yangzhe Li}, \bibinfo{person}{Yunpu Ma}, {and} \bibinfo{person}{Volker Tresp}.} \bibinfo{year}{2024}\natexlab{}.
\newblock \showarticletitle{GenTKG: Generative Forecasting on Temporal Knowledge Graph with Large Language Models}. In \bibinfo{booktitle}{\emph{Findings of the Association for Computational Linguistics: NAACL 2024}}. \bibinfo{pages}{4303--4317}.
\newblock


\bibitem[Liu et~al\mbox{.}(2021)]%
        {INDIGO}
\bibfield{author}{\bibinfo{person}{Shuwen Liu}, \bibinfo{person}{Bernardo Grau}, \bibinfo{person}{Ian Horrocks}, {and} \bibinfo{person}{Egor Kostylev}.} \bibinfo{year}{2021}\natexlab{}.
\newblock \showarticletitle{Indigo: Gnn-based inductive knowledge graph completion using pair-wise encoding}.
\newblock \bibinfo{journal}{\emph{Advances in Neural Information Processing Systems}}  \bibinfo{volume}{34} (\bibinfo{year}{2021}), \bibinfo{pages}{2034--2045}.
\newblock


\bibitem[Liu et~al\mbox{.}(2022)]%
        {TLogic}
\bibfield{author}{\bibinfo{person}{Yushan Liu}, \bibinfo{person}{Yunpu Ma}, \bibinfo{person}{Marcel Hildebrandt}, \bibinfo{person}{Mitchell Joblin}, {and} \bibinfo{person}{Volker Tresp}.} \bibinfo{year}{2022}\natexlab{}.
\newblock \showarticletitle{Tlogic: Temporal logical rules for explainable link forecasting on temporal knowledge graphs}. In \bibinfo{booktitle}{\emph{Proceedings of the AAAI conference on artificial intelligence}}, Vol.~\bibinfo{volume}{36}. \bibinfo{pages}{4120--4127}.
\newblock


\bibitem[Mei et~al\mbox{.}(2022)]%
        {ALRE-IR}
\bibfield{author}{\bibinfo{person}{Xin Mei}, \bibinfo{person}{Libin Yang}, \bibinfo{person}{Xiaoyan Cai}, {and} \bibinfo{person}{Zuowei Jiang}.} \bibinfo{year}{2022}\natexlab{}.
\newblock \showarticletitle{An Adaptive Logical Rule Embedding Model for Inductive Reasoning over Temporal Knowledge Graphs}. In \bibinfo{booktitle}{\emph{Proceedings of the 2022 Conference on Empirical Methods in Natural Language Processing}}. \bibinfo{publisher}{Association for Computational Linguistics}, \bibinfo{address}{Abu Dhabi, United Arab Emirates}, \bibinfo{pages}{7304--7316}.
\newblock
\href{https://doi.org/10.18653/v1/2022.emnlp-main.493}{doi:\nolinkurl{10.18653/v1/2022.emnlp-main.493}}


\bibitem[Mingcong et~al\mbox{.}(2024)]%
        {mgesl}
\bibfield{author}{\bibinfo{person}{Shi Mingcong}, \bibinfo{person}{Chunjiang Zhu}, \bibinfo{person}{Detian Zhang}, \bibinfo{person}{Shiting Wen}, {and} \bibinfo{person}{Li Qing}.} \bibinfo{year}{2024}\natexlab{}.
\newblock \showarticletitle{Multi-Granularity History and Entity Similarity Learning for Temporal Knowledge Graph Reasoning}. In \bibinfo{booktitle}{\emph{Proceedings of the 2024 Conference on Empirical Methods in Natural Language Processing}}. \bibinfo{pages}{5232--5243}.
\newblock


\bibitem[Pan et~al\mbox{.}(2025)]%
        {POSTRA}
\bibfield{author}{\bibinfo{person}{Jiaxin Pan}, \bibinfo{person}{Mojtaba Nayyeri}, \bibinfo{person}{Osama Mohammed}, \bibinfo{person}{Daniel Hernandez}, \bibinfo{person}{Rongchuan Zhang}, \bibinfo{person}{Cheng Cheng}, {and} \bibinfo{person}{Steffen Staab}.} \bibinfo{year}{2025}\natexlab{}.
\newblock \showarticletitle{Towards Foundation Model on Temporal Knowledge Graph Reasoning}.
\newblock \bibinfo{journal}{\emph{arXiv preprint arXiv:2506.06367}} (\bibinfo{year}{2025}).
\newblock


\bibitem[Qian et~al\mbox{.}(2025)]%
        {MLEMKD}
\bibfield{author}{\bibinfo{person}{Ye Qian}, \bibinfo{person}{Xiaoyan Wang}, \bibinfo{person}{Fuhui Sun}, {and} \bibinfo{person}{Li Pan}.} \bibinfo{year}{2025}\natexlab{}.
\newblock \showarticletitle{Compressing transfer: Mutual learning-empowered knowledge distillation for temporal knowledge graph reasoning}.
\newblock \bibinfo{journal}{\emph{IEEE Transactions on Neural Networks and Learning Systems}} (\bibinfo{year}{2025}).
\newblock


\bibitem[Rossi et~al\mbox{.}(2020)]%
        {rossi2020tgn}
\bibfield{author}{\bibinfo{person}{Emanuele Rossi}, \bibinfo{person}{Ben Chamberlain}, \bibinfo{person}{Fabrizio Frasca}, \bibinfo{person}{Davide Eynard}, \bibinfo{person}{Federico Monti}, {and} \bibinfo{person}{Michael Bronstein}.} \bibinfo{year}{2020}\natexlab{}.
\newblock \showarticletitle{Temporal Graph Networks for Deep Learning on Dynamic Graphs}. In \bibinfo{booktitle}{\emph{ICML 2020 Workshop on Graph Representation Learning and Beyond}}.
\newblock


\bibitem[Shang et~al\mbox{.}(2019)]%
        {convtranse}
\bibfield{author}{\bibinfo{person}{Chao Shang}, \bibinfo{person}{Yun Tang}, \bibinfo{person}{Jing Huang}, \bibinfo{person}{Jinbo Bi}, \bibinfo{person}{Xiaodong He}, {and} \bibinfo{person}{Bowen Zhou}.} \bibinfo{year}{2019}\natexlab{}.
\newblock \showarticletitle{End-to-end structure-aware convolutional networks for knowledge base completion}. In \bibinfo{booktitle}{\emph{Proceedings of the AAAI conference on artificial intelligence}}, Vol.~\bibinfo{volume}{33}. \bibinfo{pages}{3060--3067}.
\newblock


\bibitem[Teru et~al\mbox{.}(2020)]%
        {Grail}
\bibfield{author}{\bibinfo{person}{Komal Teru}, \bibinfo{person}{Etienne Denis}, {and} \bibinfo{person}{Will Hamilton}.} \bibinfo{year}{2020}\natexlab{}.
\newblock \showarticletitle{Inductive relation prediction by subgraph reasoning}. In \bibinfo{booktitle}{\emph{International conference on machine learning}}. PMLR, \bibinfo{pages}{9448--9457}.
\newblock


\bibitem[Trivedi et~al\mbox{.}(2017)]%
        {trivedi2017know}
\bibfield{author}{\bibinfo{person}{Rakshit Trivedi}, \bibinfo{person}{Hanjun Dai}, \bibinfo{person}{Yichen Wang}, {and} \bibinfo{person}{Le Song}.} \bibinfo{year}{2017}\natexlab{}.
\newblock \showarticletitle{Know-evolve: Deep temporal reasoning for dynamic knowledge graphs}. In \bibinfo{booktitle}{\emph{International conference on machine learning}}. PMLR, \bibinfo{pages}{3462--3471}.
\newblock


\bibitem[Trivedi et~al\mbox{.}(2019)]%
        {trivedi2019dyrep}
\bibfield{author}{\bibinfo{person}{Rakshit Trivedi}, \bibinfo{person}{Mehrdad Farajtabar}, \bibinfo{person}{Prasenjeet Biswal}, {and} \bibinfo{person}{Hongyuan Zha}.} \bibinfo{year}{2019}\natexlab{}.
\newblock \showarticletitle{{DyRep}: Learning Representations over Dynamic Graphs}. In \bibinfo{booktitle}{\emph{International Conference on Learning Representations (ICLR)}}.
\newblock


\bibitem[Vashishth et~al\mbox{.}(2020)]%
        {CompGCN}
\bibfield{author}{\bibinfo{person}{Shikhar Vashishth}, \bibinfo{person}{Soumya Sanyal}, \bibinfo{person}{Vikram Nitin}, {and} \bibinfo{person}{Partha~P Talukdar}.} \bibinfo{year}{2020}\natexlab{}.
\newblock \showarticletitle{Composition-based Multi-Relational Graph Convolutional Networks}. In \bibinfo{booktitle}{\emph{ICLR}}.
\newblock


\bibitem[Wang et~al\mbox{.}(2024)]%
        {LLM-DA}
\bibfield{author}{\bibinfo{person}{Jiapu Wang}, \bibinfo{person}{Sun Kai}, \bibinfo{person}{Linhao Luo}, \bibinfo{person}{Wei Wei}, \bibinfo{person}{Yongli Hu}, \bibinfo{person}{Alan Wee-Chung Liew}, \bibinfo{person}{Shirui Pan}, {and} \bibinfo{person}{Baocai Yin}.} \bibinfo{year}{2024}\natexlab{}.
\newblock \showarticletitle{Large language models-guided dynamic adaptation for temporal knowledge graph reasoning}.
\newblock \bibinfo{journal}{\emph{Advances in Neural Information Processing Systems}}  \bibinfo{volume}{37} (\bibinfo{year}{2024}), \bibinfo{pages}{8384--8410}.
\newblock


\bibitem[Xiong et~al\mbox{.}(2023)]%
        {Tilp}
\bibfield{author}{\bibinfo{person}{Siheng Xiong}, \bibinfo{person}{Yuan Yang}, \bibinfo{person}{Faramarz Fekri}, {and} \bibinfo{person}{James~Clayton Kerce}.} \bibinfo{year}{2023}\natexlab{}.
\newblock \showarticletitle{TILP: Differentiable Learning of Temporal Logical Rules on Knowledge Graphs}. In \bibinfo{booktitle}{\emph{The Eleventh International Conference on Learning Representations (ICLR)}}.
\newblock


\bibitem[Xiong et~al\mbox{.}(2024)]%
        {TEILP}
\bibfield{author}{\bibinfo{person}{Siheng Xiong}, \bibinfo{person}{Yuan Yang}, \bibinfo{person}{Ali Payani}, \bibinfo{person}{James~C Kerce}, {and} \bibinfo{person}{Faramarz Fekri}.} \bibinfo{year}{2024}\natexlab{}.
\newblock \showarticletitle{Teilp: Time prediction over knowledge graphs via logical reasoning}. In \bibinfo{booktitle}{\emph{Proceedings of the AAAI conference on artificial intelligence}}, Vol.~\bibinfo{volume}{38}. \bibinfo{pages}{16112--16119}.
\newblock


\bibitem[Xu et~al\mbox{.}(2020)]%
        {xu2020tgat}
\bibfield{author}{\bibinfo{person}{Da Xu}, \bibinfo{person}{Chuanwei Ruan}, \bibinfo{person}{Evren Korpeoglu}, \bibinfo{person}{Sushant Kumar}, {and} \bibinfo{person}{Kannan Achan}.} \bibinfo{year}{2020}\natexlab{}.
\newblock \showarticletitle{Inductive Representation Learning on Temporal Graphs}. In \bibinfo{booktitle}{\emph{International Conference on Learning Representations (ICLR)}}.
\newblock


\bibitem[Xu et~al\mbox{.}(2023a)]%
        {PPT}
\bibfield{author}{\bibinfo{person}{Wenjie Xu}, \bibinfo{person}{Ben Liu}, \bibinfo{person}{Miao Peng}, \bibinfo{person}{Xu Jia}, {and} \bibinfo{person}{Min Peng}.} \bibinfo{year}{2023}\natexlab{a}.
\newblock \showarticletitle{Pre-trained Language Model with Prompts for Temporal Knowledge Graph Completion}. In \bibinfo{booktitle}{\emph{Findings of the Association for Computational Linguistics: ACL 2023}}. \bibinfo{pages}{7790--7803}.
\newblock


\bibitem[Xu et~al\mbox{.}(2023b)]%
        {CENT}
\bibfield{author}{\bibinfo{person}{Yi Xu}, \bibinfo{person}{Junjie Ou}, \bibinfo{person}{Hui Xu}, {and} \bibinfo{person}{Luoyi Fu}.} \bibinfo{year}{2023}\natexlab{b}.
\newblock \showarticletitle{Temporal Knowledge Graph Reasoning with Historical Contrastive Learning}. In \bibinfo{booktitle}{\emph{Thirty-Seventh {AAAI} Conference on Artificial Intelligence, {AAAI} 2023, Thirty-Fifth Conference on Innovative Applications of Artificial Intelligence, {IAAI} 2023, Thirteenth Symposium on Educational Advances in Artificial Intelligence, {EAAI} 2023, Washington, DC, USA, February 7-14, 2023}}, \bibfield{editor}{\bibinfo{person}{Brian Williams}, \bibinfo{person}{Yiling Chen}, {and} \bibinfo{person}{Jennifer Neville}} (Eds.). \bibinfo{publisher}{{AAAI} Press}, \bibinfo{pages}{4765--4773}.
\newblock
\href{https://doi.org/10.1609/AAAI.V37I4.25601}{doi:\nolinkurl{10.1609/AAAI.V37I4.25601}}


\bibitem[Zbontar et~al\mbox{.}(2021)]%
        {collapse}
\bibfield{author}{\bibinfo{person}{Jure Zbontar}, \bibinfo{person}{Li Jing}, \bibinfo{person}{Ishan Misra}, \bibinfo{person}{Yann LeCun}, {and} \bibinfo{person}{St{\'e}phane Deny}.} \bibinfo{year}{2021}\natexlab{}.
\newblock \showarticletitle{Barlow twins: Self-supervised learning via redundancy reduction}. In \bibinfo{booktitle}{\emph{International conference on machine learning}}. PMLR, \bibinfo{pages}{12310--12320}.
\newblock


\bibitem[Zhang et~al\mbox{.}(2025)]%
        {HisRes}
\bibfield{author}{\bibinfo{person}{Jinchuan Zhang}, \bibinfo{person}{Ming Sun}, \bibinfo{person}{Chong Mu}, \bibinfo{person}{Jinhao Zhang}, \bibinfo{person}{Quanjiang Guo}, {and} \bibinfo{person}{Ling Tian}.} \bibinfo{year}{2025}\natexlab{}.
\newblock \showarticletitle{Historically relevant event structuring for temporal knowledge graph reasoning}. In \bibinfo{booktitle}{\emph{2025 IEEE 41st International Conference on Data Engineering (ICDE)}}. IEEE, \bibinfo{pages}{3179--3192}.
\newblock


\bibitem[Zhao et~al\mbox{.}(2026)]%
        {transfir}
\bibfield{author}{\bibinfo{person}{Ze Zhao}, \bibinfo{person}{Yuhui He}, \bibinfo{person}{Lyuwen Wu}, \bibinfo{person}{Gu Tang}, \bibinfo{person}{Bin Lu}, \bibinfo{person}{Xiaoying Gan}, \bibinfo{person}{Luoyi Fu}, \bibinfo{person}{Xinbing Wang}, {and} \bibinfo{person}{Chenghu Zhou}.} \bibinfo{year}{2026}\natexlab{}.
\newblock \showarticletitle{Inductive Reasoning for Temporal Knowledge Graphs with Emerging Entities}. In \bibinfo{booktitle}{\emph{The Fourteenth International Conference on Learning Representations (ICLR)}}.
\newblock


\bibitem[Zheng et~al\mbox{.}(2025)]%
        {dynamic_graph}
\bibfield{author}{\bibinfo{person}{Yanping Zheng}, \bibinfo{person}{Lu Yi}, {and} \bibinfo{person}{Zhewei Wei}.} \bibinfo{year}{2025}\natexlab{}.
\newblock \showarticletitle{A survey of dynamic graph neural networks}.
\newblock \bibinfo{journal}{\emph{Frontiers of Computer Science}} \bibinfo{volume}{19}, \bibinfo{number}{6} (\bibinfo{year}{2025}), \bibinfo{pages}{196323}.
\newblock


\bibitem[Zhu et~al\mbox{.}(2021)]%
        {CyGNet}
\bibfield{author}{\bibinfo{person}{Cunchao Zhu}, \bibinfo{person}{Muhao Chen}, \bibinfo{person}{Changjun Fan}, \bibinfo{person}{Guangquan Cheng}, {and} \bibinfo{person}{Yan Zhang}.} \bibinfo{year}{2021}\natexlab{}.
\newblock \showarticletitle{Learning from history: Modeling temporal knowledge graphs with sequential copy-generation networks}. In \bibinfo{booktitle}{\emph{Proceedings of the AAAI conference on artificial intelligence}}, Vol.~\bibinfo{volume}{35}. \bibinfo{pages}{4732--4740}.
\newblock


\end{thebibliography}

\FloatBarrier
\clearpage
\appendix
\renewcommand{\thefigure}{\thesection.\arabic{figure}}
\renewcommand{\thetable}{\thesection.\arabic{table}}
\counterwithin{figure}{section}
\counterwithin{table}{section}
\numberwithin{equation}{section}

\section{Dataset Details}
\label{app:dataset}

\textbf{Dataset statistics.}
We conduct experiments on four temporal knowledge graph benchmarks: three derived from the ICEWS event corpus~\cite{DVN_28075_2015} (\textbf{ICEWS14}, \textbf{ICEWS18}, \textbf{ICEWS05-15}) and \textbf{GDELT}~\cite{leetaru2013gdelt}.
The ICEWS benchmarks record timestamped geopolitical events at a daily granularity, while GDELT records global events at a 15-minute granularity, and all four share the extrapolation protocol commonly used in prior TKG work~\cite{REGCN,LogCL,transfir}.
Table~\ref{tab:dataset_stats} summarizes the per-benchmark statistics.
In the table, \textit{\#Snapshots} is the number of distinct timestamps in the graph and \textit{\#Emerging} is the number of entities that first appear in the validation or test split with no training-time interactions.
To enable a direct comparison with TransFIR~\cite{transfir}, we adopt the same chronological train/validation/test split with ratio 5:2:3, so that emerging-entity queries remain a meaningful fraction of the evaluation.
\begin{table}[h]
\centering
\caption{Statistics of the four TKG benchmarks used in our experiments.}
\label{tab:dataset_stats}
\adjustbox{max width=\linewidth}{
\small
\begin{tabular}{lccccc}
\toprule
\textbf{Dataset} & \textbf{\#Entities} & \textbf{\#Relations} & \textbf{\#Snapshots} & \textbf{\#Total Triples} & \textbf{\#Emerging Entities} \\
\midrule
ICEWS14     & 7{,}128  & 230 & 365     & 90{,}730   & 1{,}301 \\
ICEWS18     & 23{,}033 & 256 & 304     & 468{,}558  & 3{,}434 \\
ICEWS05-15  & 10{,}488 & 251 & 4{,}017 & 461{,}329  & 1{,}954 \\
GDELT       & 7{,}691  & 240 & 2{,}976 & 2{,}277{,}405 & 875 \\
\bottomrule
\end{tabular}
}
\end{table}

\textbf{Data splits and preprocessing.}
We adopt the chronological train / validation / test split released with each benchmark, ensuring that no future information leaks into training.
Following common practice~\cite{LogCL,transfir}, we augment each fact $(e_s,r,e_o,t)$ with its inverse $(e_o,r^{-1},e_s,t)$ and evaluate both directions, reporting the average across the two directions.
For entities that possess textual descriptions, we use the officially released entity-name strings as input to the text encoder described in Appendix~\ref{app:implementation}, while entities without descriptions receive a special placeholder token.

\textbf{Emerging-entity statistics.}
The \#Emerging column of Table~\ref{tab:dataset_stats} reports the number of entities that appear for the first time only in the validation or test split (i.e., having no training-time interaction). Across the four benchmarks, emerging entities account for a non-trivial fraction of unique entities, which justifies treating the \textit{Emerging} slice as the primary evaluation regime.

\vspace{25pt}
\section{Baselines}
\label{app:baselines}

We adopt the same baseline set as TransFIR~\cite{transfir}, comprising prior TKG reasoning methods grouped into three categories (graph-based, path-based, and static inductive), with TransFIR itself included as our direct baseline within the static-inductive group.
Short descriptions of each method are provided below.
Implementation sources follow the official releases when available, otherwise the reimplementations used in TransFIR for fair comparability.

\textbf{Graph-based.}
\begin{itemize}[leftmargin=0.6em, itemsep=0.0pt, topsep=0.0pt]
\item \textbf{CyGNet}~\cite{CyGNet}: a sequential copy-generation network that predicts future facts from recurrent historical patterns under a closed entity vocabulary.
\item \textbf{RE-GCN}~\cite{REGCN}: a recurrent evolutional GCN that learns time-evolving entity and relation representations from adjacent snapshots.
\item \textbf{HiSMatch}~\cite{Hismatch}: a historical-structure matching model that aligns query- and candidate-side historical subgraphs for link prediction.
\item \textbf{MGESL}~\cite{mgesl}: a multi-granularity history and entity similarity learning framework that captures temporal patterns at multiple levels of abstraction.
\item \textbf{LogCL}~\cite{LogCL}: a local-global history-aware contrastive-learning model that combines entity-aware attention with granularity-aware contrastive signals.
\item \textbf{HisRes}~\cite{HisRes}: a historically relevant event structuring framework with multi-granularity evolutionary and global relevance encoders.
\item \textbf{MLEMKD}~\cite{MLEMKD}: a mutual learning-empowered knowledge distillation method that compresses a TKG reasoner through adaptive distillation.
\end{itemize}

\textbf{Path-based.}
\begin{itemize}[leftmargin=0.6em, itemsep=0.0pt, topsep=0.0pt]
\item \textbf{TLogic}~\cite{TLogic}: an explainable temporal-rule forecasting framework that extracts cyclic rules via time-constrained random walks.
\item \textbf{TILP}~\cite{Tilp}: a differentiable learning framework for temporal logical rules on knowledge graphs.
\item \textbf{ECEformer}~\cite{ECEformer}: a Transformer that encodes evolutionary chains of events through intra-quadruple representation learning and inter-quadruple context mixing.
\item \textbf{GenTKG}~\cite{GenTKG}: a retrieval-augmented generation framework that combines temporal logical-rule retrieval with few-shot instruction tuning of a large language model.
\end{itemize}

\textbf{Static inductive.}
\begin{itemize}[leftmargin=0.6em, itemsep=0.0pt, topsep=0.0pt]
\item \textbf{CompGCN}~\cite{CompGCN}: a composition-based multi-relational GCN that unifies node and relation embeddings under a shared message-passing framework.
\item \textbf{ICL}~\cite{ICL}: an in-context-learning approach that performs temporal KG forecasting by prompting a frozen large language model with historical quadruples.
\item \textbf{PPT}~\cite{PPT}: a pretrained-language-model method that reformulates TKG completion as a cloze-style task with soft prompts.
\item \textbf{MorsE}~\cite{morse}: a meta-knowledge transfer framework that learns entity-independent structural patterns for inductive KG embedding of unseen entities.
\item \textbf{InGram}~\cite{InGram}: an inductive KG embedding model that exploits a learned relation graph to generalize to unseen entities and relations.
\item \textbf{TransFIR}~\cite{transfir}: our direct baseline and the current state of the art on the emerging slice, which maps each entity to a learned codebook and transfers type-level behavioral prototypes from semantically similar cluster members.
\end{itemize}

\vspace{25pt}
\section{Implementation Details}
\label{app:implementation}

\textbf{Static encoder.}
For every entity $e\!\in\!\mathcal{E}$, we obtain a frozen textual representation $\mathbf{h}_e$ by feeding the entity surface form to a pretrained BERT-base encoder~\cite{bert} and taking the \texttt{[CLS]} token, with the encoder weights not updated during training.
For a direct comparison with TransFIR~\cite{transfir}, we use the same encoder choice.

\textbf{Relational decoder.}
We score candidate triples with ConvTransE~\cite{convtranse}, which has been widely adopted by recent TKG reasoning methods~\cite{REGCN,LogCL,transfir}.
The decoder's kernel size, number of filters, and dropout are kept at their default values from the public ConvTransE implementation.

\textbf{Hyperparameter grid.}
We select hyperparameters by grid search on the validation MRR of the \textit{Emerging} slice, which is the setting most directly aligned with our research goal.
Table~\ref{tab:hp_grid} lists the search ranges shared with TransFIR~\cite{transfir} and the configuration selected per dataset for AdaTKG-EMA (default operator).
The EMA decay rate $\alpha\!=\!\sigma(\rho)$ is a learnable scalar (Section~\ref{subsec:adatkg}) and therefore is not searched over.

\begin{table}[h]
\centering
\caption{Hyperparameter search ranges and the selected configuration for AdaTKG-EMA.
The ranges for shared hyperparameters (chain length, GNN layers, hidden dimensionality, codebook size) match those of TransFIR~\cite{transfir} for a direct comparison.}
\label{tab:hp_grid}
\adjustbox{max width=\linewidth}{
\small
\begin{tabular}{lc|cccc}
\toprule
\multirow{2.5}{*}{\textbf{Hyperparameter}} & \multirow{2.5}{*}{\textbf{Search Range}}
  & \multicolumn{4}{c}{\textbf{Selected (AdaTKG-EMA)}} \\
\cmidrule(lr){3-6}
                              &                                & ICEWS14 & ICEWS18 & ICEWS05-15 & GDELT \\
\midrule
Interaction-chain length $L$  & $\{10,\,15,\,30\}$             & 10 & 30 & 30 & 30 \\
Number of layers              & $\{2,\,3\}$                    & 3  & 3  & 3  & 2  \\
Hidden dimensionality $d$     & $\{256,\,512,\,768,\,1024\}$   & 768 & 1024 & 1024 & 1024 \\
Codebook size $K$             & $\{30,\,50,\,100\}$            & 100 & 100 & 50 & 30 \\
\bottomrule
\end{tabular}
}
\end{table}

\textbf{Training objective.}
We follow the loss formulation of TransFIR~\cite{transfir} without modification, and our only design choice is the absence of any auxiliary loss on the memory state or the gate, so that both modules are trained purely through the link-prediction signal.
AdaTKG is optimized end-to-end under the cross-entropy loss and a vector-quantization commitment loss applied to the codebook as
\begin{equation}
\label{eq:loss}
\mathcal{L} \;=\; \mathcal{L}_{\mathrm{LP}} \;+\; \lambda\,\mathcal{L}_{\mathrm{VQ}},
\qquad \lambda = 0.1.
\end{equation}

\textbf{Training protocol.}
All models are optimized with Adam~\cite{kingma2014adam}.
We train for up to 200 epochs with early stopping (patience of 10 epochs on validation MRR) and report the best checkpoint.
Each experiment is repeated with three random seeds and we report the mean across seeds in the main paper, with standard deviations relegated to the Additional Experiments appendix.

\textbf{Hardware and runtime.}
All experiments are conducted on a single NVIDIA L40S GPU (48\,GB).
A full training run of AdaTKG-EMA on ICEWS14 at the selected hyperparameters of Table~\ref{tab:hp_grid} takes about $1$ hour, with peak GPU memory under $30$\,GB.
Training time on the larger benchmarks (ICEWS18, ICEWS05-15, GDELT) is reported per-epoch in the efficiency tables of Appendix~\ref{app:efficiency_others}.

\vspace{25pt}
\section{Backbone Details}
\label{app:backbone}

For completeness we reproduce the three backbone outputs $\textsc{Backbone}(q)\!\to\!(\mathbf{h}_e,\,\omega_e\!\cdot\!\mathbf{c}_{\pi(e),t},\,\mathbf{x}^{(\tau)}_e)$ that AdaTKG consumes (Section~\ref{subsec:adatkg}\,[1]).
All three modules are taken from~\cite{transfir} unchanged.

\textbf{(a) Static entity embedding $\mathbf{h}_e$.}
$\mathbf{h}_e\!\in\!\mathbb{R}^d$ is obtained by feeding the textual surface form of entity $e$ into a frozen pretrained BERT~\cite{bert} encoder and reading the \texttt{[CLS]} token. Because the encoder is not fine-tuned, $\mathbf{h}_e$ is well-defined for every entity that has a textual description --- in particular, for entities never seen at training time --- which is the source of the model's inductivity.

\textbf{(b) Type-level inductive prior $\omega_e\!\cdot\!\mathbf{c}_{\pi(e),t}$.}
A learnable vector-quantized codebook $\mathcal{C}\!=\!\{\mathbf{c}_k\}_{k=1}^{K}$ is applied to $\mathbf{h}_e$ to produce a cluster assignment $\pi(e)\!=\!\arg\min_k\|\mathbf{h}_e-\mathbf{c}_k\|_2^2$. At every query timestamp $t$, the cluster prototype $\mathbf{c}_{\pi(e),t}$ is recomputed by pooling the IC-encoder outputs of all cluster mates, so the prototype evolves as the graph evolves. The transfer gate $\omega_e\!=\!\Psi([\mathbf{h}_e\,\|\,\mathbf{c}_{\pi(e),t}])\!\in\![0,1]^d$ regulates how much of the prototype each entity inherits. The codebook is trained jointly with the rest of the model through the commitment loss $\mathcal{L}_{\mathrm{VQ}}$ in Eq.~\eqref{eq:loss}.

\textbf{(c) Interaction signal $\mathbf{x}^{(\tau)}_e$.}
For each fact in which $e$ participates, the event is summarized as
\begin{equation}
\label{eq:interaction_encoder}
\mathbf{x}^{(\tau)}_e \;=\; \mathbf{W}_2 \,\mathrm{GELU}\!\bigl(\mathbf{W}_1\,[\mathbf{h}^{\mathrm{IC}}_e \,\|\, \mathbf{h}_{r_\tau}]\bigr),
\end{equation}
with learnable projections $\mathbf{W}_1\!\in\!\mathbb{R}^{d\times 2d}$ and $\mathbf{W}_2\!\in\!\mathbb{R}^{d\times d}$. The two inputs are
(i) $\mathbf{h}^{\mathrm{IC}}_e$, the embedding of $e$'s \emph{interaction chain} --- the time-ordered sequence of recent facts involving $e$, filtered by relation similarity to the query~\cite{transfir} and encoded with a Transformer --- and
(ii) $\mathbf{h}_{r_\tau}$, the relation embedding of the $\tau$-th observed fact involving $e$ (\emph{not} the relation of an open query).

Both inputs are known at prediction time, so $\mathbf{x}^{(\tau)}_e$ never observes the ground-truth object of an open query. At the first appearance of an emerging entity, the chain is empty and $\mathbf{m}_e\!=\!\mathbf{0}$, which the adaptive gate of Eq.~\eqref{eq:gate} converts into a hard fall-back to the static-inductive prior (Corollary~1, Appendix~\ref{app:formal_analysis}).

\vspace{25pt}
\section{Reduction of AdaTKG to the Static-Inductive Setting at Cold Start}
\label{app:formal_analysis}

In this appendix, we formalize the observation, stated informally in Section~\ref{subsec:paradigms} of the main text, that AdaTKG strictly generalizes the \textit{Static + Inductive} setting (whose representative instantiation is TransFIR~\cite{transfir}), and we make precise the sense in which AdaTKG incurs no cold-start cost relative to that setting.
The analysis concerns only the gated fusion rule, and therefore holds independently of the specific stateful update operator $\mathcal{U}$ used to maintain the per-entity memory.

\textbf{AdaTKG generalizes the static-inductive setting.}
We first show that AdaTKG recovers the static-inductive prior exactly when the adaptive gate vanishes.

\noindent\textbf{Proposition 1} (\textit{Reduction to the static-inductive setting at zero gate}).
\label{prop:reduction}
\textit{Let $\mathbf{z}^{(t_q),\,\textup{Ada}}_e$ and $\mathbf{z}^{(t_q),\,\textup{SI}}_e$ denote, respectively, the effective entity representation produced by AdaTKG via Eq.~\eqref{eq:adaptive_fusion} and by the static-inductive setting via Eq.~\eqref{eq:static_inductive} at query time $t_q$.
If the adaptive gate satisfies $g^{(t_q)}_e = \mathbf{0}$ and the static-inductive prior used by AdaTKG coincides with the one defined in Eq.~\eqref{eq:static_inductive}, then}
\[
\mathbf{z}^{(t_q),\,\textup{Ada}}_e \;=\; \mathbf{z}^{(t_q),\,\textup{SI}}_e ,
\]
\textit{and consequently the score $\phi_{t_q}(e_q,r_q,e_o)$ of AdaTKG equals that of the static-inductive setting for every query triple $(e_q,r_q,e_o)$.}

\noindent\textbf{Proof.}
Substituting $g^{(t_q)}_e = \mathbf{0}$ into the AdaTKG fusion rule of Eq.~\eqref{eq:adaptive_fusion} yields
\[
\mathbf{z}^{(t_q),\,\textup{Ada}}_e
\;=\; (1 - \mathbf{0}) \odot \bigl( \mathbf{h}_e + \omega_e \cdot \mathbf{c}_{\pi(e),\,t_q} \bigr)
\;+\; \mathbf{0} \odot \mathbf{m}_e
\;=\; \mathbf{h}_e + \omega_e \cdot \mathbf{c}_{\pi(e),\,t_q}
\;=\; \mathbf{z}^{(t_q),\,\textup{SI}}_e ,
\]
where the last equality is by the \textit{Static + Inductive} definition in Eq.~\eqref{eq:static_inductive}.
Because the relational decoder $f(\cdot)$ in the unified scoring form (Section~\ref{subsec:paradigms}) depends on entity representations only through $\mathbf{z}^{(t_q)}_e$, replacing $\mathbf{z}^{(t_q),\,\textup{Ada}}_e$ by $\mathbf{z}^{(t_q),\,\textup{SI}}_e$ yields identical scores. $\hfill\square$

\textbf{Cold-start parity.}
Proposition~\ref{prop:reduction} has an immediate consequence at the moment each entity first appears in the graph.

\noindent\textbf{Corollary 1} (\textit{Cold-start parity with the static-inductive setting}).
\textit{At the query time $t_q = t_e(e)$ of an emerging entity $e$, the per-entity memory is zero by the initialization in Eq.~\eqref{eq:memory_update}, namely $\mathbf{m}_e = \mathbf{0}$.
If the adaptive gate is parameterized so that $g^{(t_q)}_e = \mathbf{0}$ whenever $\mathbf{m}_e = \mathbf{0}$ (a design choice readily enforced by zero-masking the gate output when the memory buffer is empty), then, by Proposition~\ref{prop:reduction}, the prediction of AdaTKG coincides exactly with that of the static-inductive setting.}

Corollary~1 formalizes the graceful-degradation property emphasized in the main text: AdaTKG never pays a cold-start cost relative to the static-inductive setting (instantiated by TransFIR~\cite{transfir} in our experiments), and all empirical gains reported in Section~\ref{sec:experiments} stem purely from information retained in the per-entity memory.

\vspace{25pt}
\section{Stateful Update Operators}
\label{app:operators}

We compared three instantiations of the stateful update operator $\mathcal{U}$ in Section~\ref{subsec:ablation}: a learnable EMA (used by AdaTKG), a GRU cell, and a cross-attention readout over a bounded buffer. All three follow the unified template of Eq.~\eqref{eq:memory_update},
\begin{equation*}
\mathbf{m}^{(\tau)}_e \;=\; \mathcal{U}\!\bigl(\mathbf{m}^{(\tau-1)}_e,\,\mathbf{x}^{(\tau)}_e\bigr),
\qquad
\mathbf{m}^{(0)}_e \;=\; \mathbf{0},
\end{equation*}
and share the same interaction signal
\begin{equation}
\label{eq:interaction_signal}
\mathbf{x}^{(\tau)}_e \;=\; \mathrm{MLP}_{\mathrm{enc}}\!\bigl([\,\mathbf{c}^{(\tau)}_e \,\|\, \mathbf{r}^{(\tau)}_e\,]\bigr),
\end{equation}
where $\mathbf{c}^{(\tau)}_e$ is the chain summary produced by the TransFIR backbone at the $\tau$-th interaction of $e$ and $\mathbf{r}^{(\tau)}_e$ is the relation embedding of that interaction.
The three variants differ only in how $\mathcal{U}$ aggregates the past memory state with the new signal.

\textbf{(a) Learnable EMA (AdaTKG, Section~\ref{subsec:adatkg}).}
The default instantiation realizes $\mathcal{U}$ as an exponential moving average parameterized by a single shared scalar:
\begin{equation}
\label{eq:U_ema}
\mathbf{m}^{(\tau)}_e \;=\; \alpha\,\mathbf{m}^{(\tau-1)}_e \;+\; (1-\alpha)\,\mathbf{x}^{(\tau)}_e,
\qquad
\alpha \;=\; \sigma(\rho),\ \rho\!\in\!\mathbb{R}.
\end{equation}
The decay $\alpha$ is shared across all entities, so the operator introduces only one learnable scalar beyond TransFIR.

\textbf{(b) GRU cell.}
A standard GRU cell~\cite{cho2014gru} treats $\mathbf{m}^{(\tau-1)}_e$ as the hidden state and $\mathbf{x}^{(\tau)}_e$ as the input, applying a learned reset and update gate:
\begin{equation}
\label{eq:U_gru}
\mathbf{m}^{(\tau)}_e \;=\; \mathrm{GRUCell}\!\bigl(\mathbf{x}^{(\tau)}_e,\,\mathbf{m}^{(\tau-1)}_e\bigr).
\end{equation}
The GRU cell adds three weight matrices in $\mathbb{R}^{d\times d}$ and three biases, roughly $3d^2$ extra parameters.

\textbf{(c) Cross-attention readout over a bounded buffer.}
This variant maintains a per-entity FIFO buffer $\mathbf{B}_e\!\in\!\mathbb{R}^{K\times d}$ of the most recent $K$ interaction signals (default $K\!=\!16$). The buffer is updated by appending the new signal at position $\tau \!\!\mod\!\! K$,
\begin{equation*}
\mathbf{B}_e[\tau \bmod K] \;\gets\; \mathbf{x}^{(\tau)}_e,
\end{equation*}
and the memory state read at query time is computed by cross-attending the query embedding $\mathbf{q}_e$ to the buffer,
\begin{equation}
\label{eq:U_attn}
\mathbf{m}^{(\tau)}_e \;=\; \mathrm{MultiHeadAttn}\!\bigl(\mathbf{q}_e,\,\mathbf{B}_e,\,\mathbf{B}_e\bigr).
\end{equation}
This variant adds $\mathcal{O}(d^2)$ attention parameters and an additional $\mathcal{O}(K\,d)$ buffer per entity.

\textbf{Adaptive gate fusion (shared).}
Regardless of the choice of $\mathcal{U}$, the resulting memory $\mathbf{m}_e$ is fused with the inductive prior through the same adaptive gate of Eq.~\eqref{eq:gate}:
\begin{equation*}
\mathbf{z}^{(t_q)}_e
\;=\;
(1-g^{(t_q)}_e)\odot(\mathbf{h}_e+\omega_e\,\mathbf{c}_{\pi(e),\,t_q})
\;+\;
g^{(t_q)}_e\odot\mathbf{m}_e,
\end{equation*}
with the cold-start zero-mask $g^{(t_q)}_e\!=\!\mathbf{0}$ active whenever $\mathbf{m}_e\!=\!\mathbf{0}$.
The (P1)--(P3) protocol of Section~\ref{subsec:adatkg} (epoch reset, chronological replay, detached update) is also applied identically to all three variants, so the comparison in Section~\ref{subsec:ablation} isolates the effect of $\mathcal{U}$ alone.

\textbf{Parameter count summary.}
\begin{itemize}[leftmargin=0.6em, itemsep=0.0pt, topsep=0.0pt]
\item EMA: $1$ scalar (shared $\rho$).
\item GRU: $\sim\!3d^2$ parameters in the cell.
\item Attention: $\sim\!4d^2$ attention parameters and a per-entity buffer of size $K\,d$.
\end{itemize}
The single-scalar parameterization of EMA is the most parsimonious operating point along this frontier and, as reported in Section~\ref{subsec:ablation}, also the most accurate on the emerging slice.

\vspace{25pt}
\section{Sensitivity to Memory Update Timing}
\label{app:update_timing}

By default, AdaTKG updates the per-entity memory $\mathbf{m}_e$ with the current interaction signal $\mathbf{x}^{(\tau)}_e$ \emph{before} reading $\mathbf{m}_e$ for scoring (the \texttt{before} ordering used throughout the main paper). 
To probe the sensitivity of this design choice, we additionally train AdaTKG with the alternative \texttt{after} ordering, in which $\mathbf{m}_e$ is read for scoring \emph{first},so the score depends only on the strictly past memory and the EMA update is committed only afterwards. Both orderings share the per-dataset best HP of AdaTKG (Table~\ref{tab:hp_grid}); only the order of the read and the update differs. Table~\ref{tab:update_timing} reports the comparison on the \emph{Emerging} slice,
demonstarting that across all four benchmarks, the default \texttt{before} ordering consistently outperforms the \texttt{after} ordering, supporting our design choice in the main paper.

\begin{table}[h]
\centering
\caption{\textbf{Sensitivity to memory update timing.} AdaTKG on the \emph{Emerging} slice with the memory update applied either \texttt{before} (default) or \texttt{after} scoring.}
\label{tab:update_timing}
\adjustbox{max width=\linewidth}{
\small
\begin{tabular}{l|ccc|ccc|ccc|ccc}
\toprule
\multirow{2.5}{*}{\textbf{Update timing}}
  & \multicolumn{3}{c|}{\textbf{ICEWS14}}
  & \multicolumn{3}{c|}{\textbf{ICEWS18}}
  & \multicolumn{3}{c|}{\textbf{ICEWS05-15}}
  & \multicolumn{3}{c}{\textbf{GDELT}} \\
\cmidrule(lr){2-4}\cmidrule(lr){5-7}\cmidrule(lr){8-10}\cmidrule(lr){11-13}
  & MRR & H@3 & H@10
  & MRR & H@3 & H@10
  & MRR & H@3 & H@10
  & MRR & H@3 & H@10 \\
\midrule
\texttt{before} (default) & \first{.2011} & \first{.2250} & \first{.3621} & \first{.1379} & \first{.1543} & \first{.2612} & \first{.2270} & \first{.2573} & \first{.3850} & \first{.1051} & \first{.1129} & \first{.2301} \\
\texttt{after}            & .1777 & .2011 & .3505 & .1322 & .1470 & .2482 & .1914 & .2110 & .3290 & .0814 & .0797 & .1663 \\
\bottomrule
\end{tabular}
}
\end{table}

\vspace{20pt}
\section{Multi-Seed Robustness}
\label{app:multiseed}
To ensure fair comparison across all baselines, the main paper reports single-seed results. In this section, we further evaluate AdaTKG and the static-inductive baseline (\cite{transfir}) under three random seeds at the per-dataset best HP of Table~\ref{tab:hp_grid}, reporting the mean and standard deviation on the \emph{Emerging} slice (Table~\ref{tab:multiseed}). AdaTKG continues to outperform the SoTA under this multi-seed setting.

\begin{table}[h]
\centering
\caption{\textbf{Multi-seed robustness on the \emph{Emerging} slice.} Mean $\pm$ standard deviation over three random seeds.
}
\label{tab:multiseed}
\adjustbox{max width=\linewidth}{
\footnotesize
\begin{tabular}{ll|ccc}
\toprule
\multirow{2.5}{*}{\textbf{Dataset}} & \multirow{2.5}{*}{\textbf{Metric}}
  & \multicolumn{3}{c}{\textbf{Method}} \\
\cmidrule(lr){3-5}
  & & TransFIR~\cite{transfir} & \cellcolor{yellow!15}\textbf{AdaTKG (Ours)} & \textit{$\Delta$ (\%)} \\
\midrule
\multirow{3}{*}{ICEWS14}
  & MRR  & .1768 $\pm$ .0012 & \cellcolor{yellow!15} \first{.2008} $\pm$ .0014 & \textbf{\up{+13.6}} \\
  & H@3  & .2090 $\pm$ .0015 & \cellcolor{yellow!15} \first{.2256} $\pm$ .0017 & \textbf{\up{+7.9}}  \\
  & H@10 & .3417 $\pm$ .0019 & \cellcolor{yellow!15} \first{.3618} $\pm$ .0021 & \textbf{\up{+5.9}}  \\
\midrule
\multirow{3}{*}{ICEWS18}
  & MRR  & .1118 $\pm$ .0011 & \cellcolor{yellow!15} \first{.1382} $\pm$ .0013 & \textbf{\up{+23.6}} \\
  & H@3  & .1224 $\pm$ .0014 & \cellcolor{yellow!15} \first{.1538} $\pm$ .0016 & \textbf{\up{+25.7}} \\
  & H@10 & .2257 $\pm$ .0017 & \cellcolor{yellow!15} \first{.2607} $\pm$ .0019 & \textbf{\up{+15.5}} \\
\midrule
\multirow{3}{*}{ICEWS05-15}
  & MRR  & .2173 $\pm$ .0010 & \cellcolor{yellow!15} \first{.2274} $\pm$ .0012 & \textbf{\up{+4.6}}  \\
  & H@3  & .2534 $\pm$ .0013 & \cellcolor{yellow!15} \first{.2569} $\pm$ .0015 & \textbf{\up{+1.4}}  \\
  & H@10 & .3712 $\pm$ .0016 & \cellcolor{yellow!15} \first{.3853} $\pm$ .0018 & \textbf{\up{+3.8}}  \\
\midrule
\multirow{3}{*}{GDELT}
  & MRR  & .1009 $\pm$ .0014 & \cellcolor{yellow!15} \first{.1054} $\pm$ .0013 & \textbf{\up{+4.5}}  \\
  & H@3  & .0998 $\pm$ .0017 & \cellcolor{yellow!15} \first{.1124} $\pm$ .0016 & \textbf{\up{+12.6}} \\
  & H@10 & .2127 $\pm$ .0020 & \cellcolor{yellow!15} \first{.2305} $\pm$ .0019 & \textbf{\up{+8.4}}  \\
\bottomrule
\end{tabular}
}
\end{table}

\vspace{10pt}
\section{Efficiency Comparison Across Benchmarks}
\label{app:efficiency_others}
Tables~\ref{tab:efficiency_icews18}, \ref{tab:efficiency_icews05_15}, and \ref{tab:efficiency_gdelt} extend the main-paper Table~\ref{tab:efficiency} to ICEWS18, ICEWS05-15, and GDELT.
The pattern observed on ICEWS14 consistently holds across all benchmarks, with every update operator incurring only 
modest
overhead while improving emerging-entity performance, indicating that the cost–benefit profile stems from the \textit{adaptivity principle} itself rather than any specific operator.

\textbf{Memory footprint at inference.}
Beyond the cost metrics in the tables above, AdaTKG keeps one additional piece of state at inference: a per-entity memory buffer of size $|\mathcal{E}|\!\times\!d$.
With our largest configuration ($d\!=\!1024$ on ICEWS18, which has $23{,}033$ entities), this buffer takes about $94$\,MB, and stays under $100$\,MB on every benchmark.

\begin{table}[H]
\centering
\caption{\textbf{Efficiency comparison on ICEWS18.}}
\label{tab:efficiency_icews18}
\adjustbox{max width=\linewidth}{
\small
\begin{tabular}{c|l|cc|cc|cc|ccc}
\toprule
\multicolumn{2}{c|}{\multirow{3}{*}{\textbf{Method}}}
  & \multicolumn{6}{c|}{\textbf{[1] Efficiency}}
  & \multicolumn{3}{c}{\textbf{[2] Performance}} \\
\cmidrule(lr){3-8}\cmidrule(lr){9-11}
\multicolumn{2}{c|}{}
  & \multicolumn{2}{c|}{\textbf{\# Parameters} (M)}
  & \multicolumn{2}{c|}{\textbf{Training time} (s/epoch)}
  & \multicolumn{2}{c|}{\textbf{FLOPs} (M/query)}
  & \multirow{2.5}{*}{\textbf{MRR}} & \multirow{2.5}{*}{\textbf{H@3}} & \multirow{2.5}{*}{\textbf{H@10}} \\
\cmidrule(lr){3-4}\cmidrule(lr){5-6}\cmidrule(lr){7-8}
\multicolumn{2}{c|}{}
  & Value & $\Delta$\,(\%)
  & Value & $\Delta$\,(\%)
  & Value & $\Delta$\,(\%)
  &       &       & \\
\midrule
\multicolumn{2}{l|}{Base (w/o Adaptivity)~\cite{transfir}}
  & 72.77   & --
  & 330.7   & --
  & 1643.9  & --
  & .1114   & .1230 & .2252 \\
\midrule
\multirow{3}{*}{\textbf{AdaTKG}}
  & EMA (\textit{default})
  & 78.02   & \textbf{+7.2\%}
  & 361.3   & \textbf{+9.2\%}
  & 1654.4  & \textbf{+0.6\%}
  & \first{.1379} & \first{.1543} & \first{.2612} \\
  & GRU
  & 84.32   & \textbf{+15.9\%}
  & 378.0   & \textbf{+14.3\%}
  & 1658.8  & \textbf{+0.9\%}
  & \first{.1428} & \first{.1599} & \first{.2605} \\
  & Cross-attention
  & 82.22   & \textbf{+13.0\%}
  & 360.5   & \textbf{+9.0\%}
  & 1725.8  & \textbf{+5.0\%}
  & \first{.1454} & \first{.1712} & \first{.2761} \\
\bottomrule
\end{tabular}
}
\end{table}

\begin{table}[H]
\centering
\caption{\textbf{Efficiency comparison on ICEWS05-15.}}
\label{tab:efficiency_icews05_15}
\adjustbox{max width=\linewidth}{
\small
\begin{tabular}{c|l|cc|cc|cc|ccc}
\toprule
\multicolumn{2}{c|}{\multirow{3}{*}{\textbf{Method}}}
  & \multicolumn{6}{c|}{\textbf{[1] Efficiency}}
  & \multicolumn{3}{c}{\textbf{[2] Performance}} \\
\cmidrule(lr){3-8}\cmidrule(lr){9-11}
\multicolumn{2}{c|}{}
  & \multicolumn{2}{c|}{\textbf{\# Parameters} (M)}
  & \multicolumn{2}{c|}{\textbf{Training time} (s/epoch)}
  & \multicolumn{2}{c|}{\textbf{FLOPs} (M/query)}
  & \multirow{2.5}{*}{\textbf{MRR}} & \multirow{2.5}{*}{\textbf{H@3}} & \multirow{2.5}{*}{\textbf{H@10}} \\
\cmidrule(lr){3-4}\cmidrule(lr){5-6}\cmidrule(lr){7-8}
\multicolumn{2}{c|}{}
  & Value & $\Delta$\,(\%)
  & Value & $\Delta$\,(\%)
  & Value & $\Delta$\,(\%)
  &       &       & \\
\midrule
\multicolumn{2}{l|}{Base (w/o Adaptivity)~\cite{transfir}}
  & 72.71   & --
  & 420.5   & --
  & 1913.9  & --
  & .2177   & .2530 & .3708 \\
\midrule
\multirow{3}{*}{\textbf{AdaTKG}}
  & EMA (\textit{default})
  & 77.96   & \textbf{+7.2\%}
  & 466.3   & \textbf{+10.9\%}
  & 1924.4  & \textbf{+0.5\%}
  & \first{.2270} & \first{.2573} & \first{.3850} \\
  & GRU
  & 84.25   & \textbf{+15.9\%}
  & 491.0   & \textbf{+16.8\%}
  & 1930.2  & \textbf{+0.8\%}
  & \first{.2330} & \first{.2700} & \first{.3925} \\
  & Cross-attention
  & 82.15   & \textbf{+13.0\%}
  & 459.8   & \textbf{+9.4\%}
  & 1995.8  & \textbf{+4.3\%}
  & \first{.2243} & \first{.2573}  & \first{.3815} \\
\bottomrule
\end{tabular}
}
\end{table}

\begin{table}[H]
\centering
\caption{\textbf{Efficiency comparison on GDELT.}}
\label{tab:efficiency_gdelt}
\adjustbox{max width=\linewidth}{
\small
\begin{tabular}{c|l|cc|cc|cc|ccc}
\toprule
\multicolumn{2}{c|}{\multirow{3}{*}{\textbf{Method}}}
  & \multicolumn{6}{c|}{\textbf{[1] Efficiency}}
  & \multicolumn{3}{c}{\textbf{[2] Performance}} \\
\cmidrule(lr){3-8}\cmidrule(lr){9-11}
\multicolumn{2}{c|}{}
  & \multicolumn{2}{c|}{\textbf{\# Parameters} (M)}
  & \multicolumn{2}{c|}{\textbf{Training time} (s/epoch)}
  & \multicolumn{2}{c|}{\textbf{FLOPs} (M/query)}
  & \multirow{2.5}{*}{\textbf{MRR}} & \multirow{2.5}{*}{\textbf{H@3}} & \multirow{2.5}{*}{\textbf{H@10}} \\
\cmidrule(lr){3-4}\cmidrule(lr){5-6}\cmidrule(lr){7-8}
\multicolumn{2}{c|}{}
  & Value & $\Delta$\,(\%)
  & Value & $\Delta$\,(\%)
  & Value & $\Delta$\,(\%)
  &       &       & \\
\midrule
\multicolumn{2}{l|}{Base (w/o Adaptivity)~\cite{transfir}}
  & 60.08   & --
  & 1384.6  & --
  & 1062.6  & --
  &        .1013  &        .0994  &        .2131  \\
\midrule
\multirow{3}{*}{\textbf{AdaTKG}}
  & EMA (\textit{default})
  & 65.33   & \textbf{+8.7\%}
  & 1488.7  & \textbf{+7.5\%}
  & 1073.1  & \textbf{+1.0\%}
  & \first{.1051} & \first{.1129} & \first{.2301} \\
  & GRU
  & 71.63   & \textbf{+19.2\%}
  & 1549.0  & \textbf{+11.9\%}
  & 1075.9  & \textbf{+1.3\%}
  & \first{.1112} & \first{.1141} & \first{.2243} \\
  & Cross-attention
  & 69.53   & \textbf{+15.7\%}
  & 1503.2  & \textbf{+8.6\%}
  & 1144.4  & \textbf{+7.7\%}
  & \first{.1297} & \first{.1396} & \first{.2544} \\
\bottomrule
\end{tabular}
}
\end{table}

\vspace{25pt}
\section{Memory Ablation Across Benchmarks}
\label{app:memory_ablation_others}
\begin{figure}[H]
\centering
\begin{adjustbox}{max width=\linewidth}
\includegraphics[width=\linewidth]{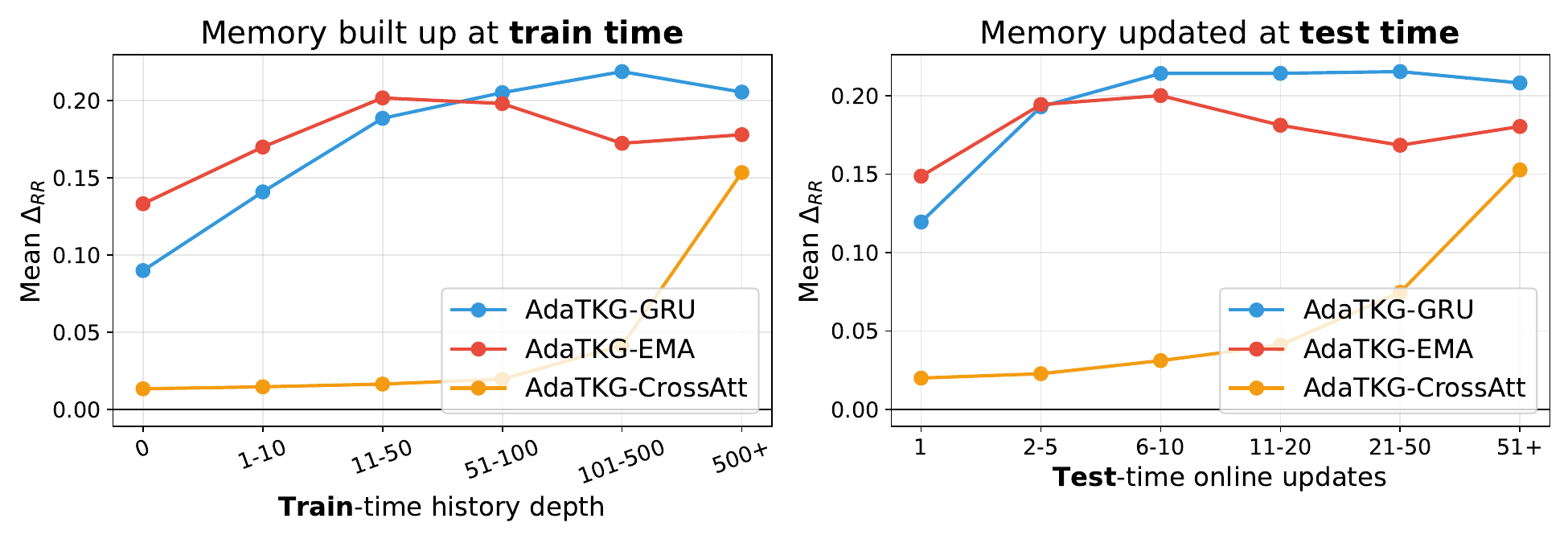}
\end{adjustbox}
\caption{\textbf{Memory ablation on ICEWS18.}}
\label{fig:memory_ablation_icews18}
\end{figure}

\begin{figure}[H]
\centering
\begin{adjustbox}{max width=\linewidth}
\includegraphics[width=\linewidth]{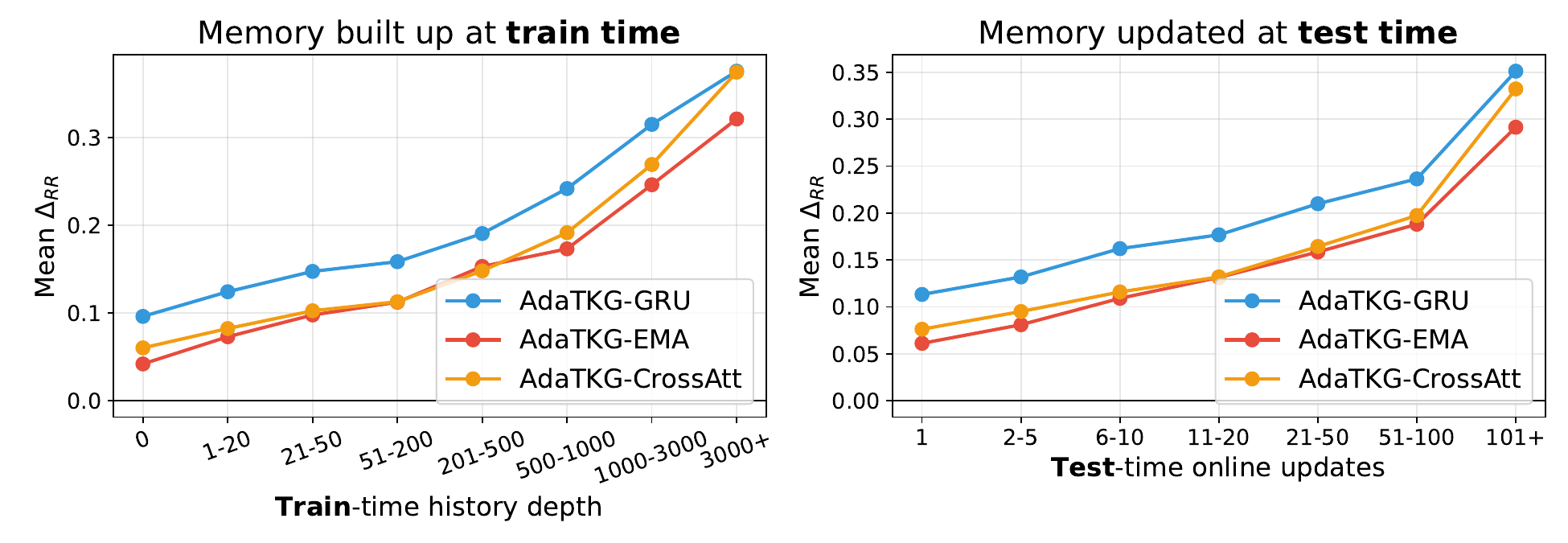}
\end{adjustbox}
\caption{\textbf{Memory ablation on ICEWS05-15.}}
\label{fig:memory_ablation_icews05_15}
\end{figure}

\begin{figure}[H]
\centering
\begin{adjustbox}{max width=\linewidth}
\includegraphics[width=\linewidth]{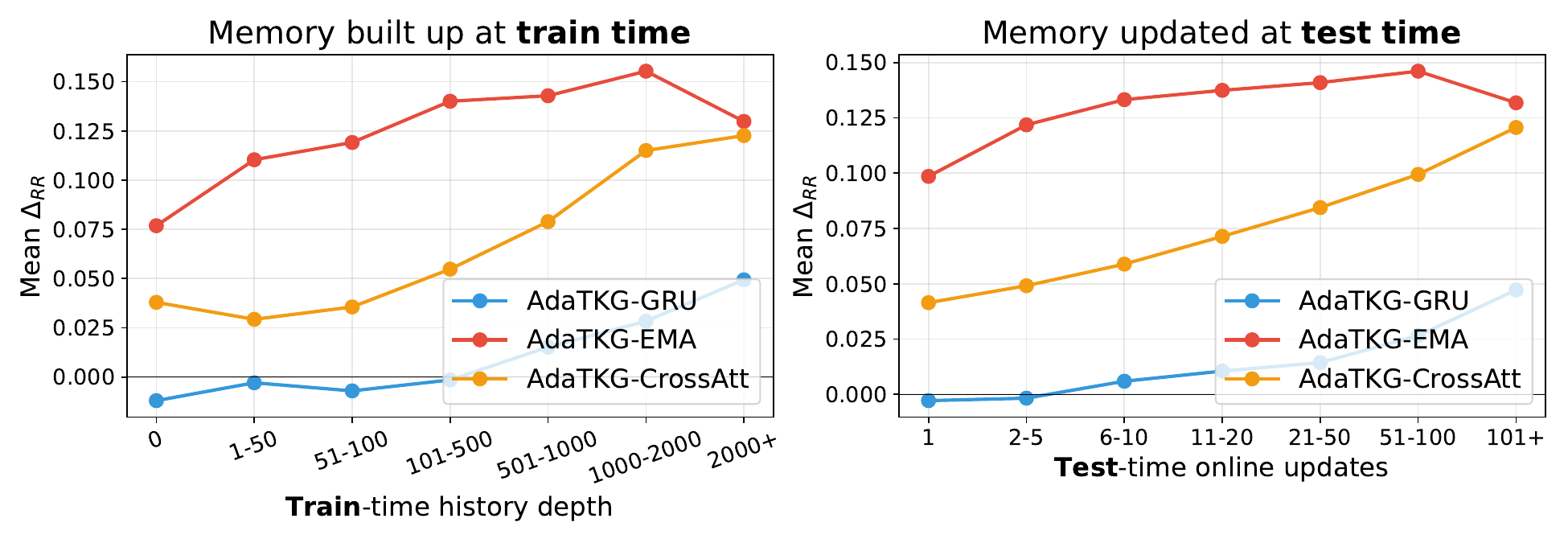}
\end{adjustbox}
\caption{\textbf{Memory ablation on GDELT.}}
\label{fig:memory_ablation_gdelt}
\end{figure}

Figures~\ref{fig:memory_ablation_icews18},~\ref{fig:memory_ablation_icews05_15}, and~\ref{fig:memory_ablation_gdelt} extend the main-paper Figure~\ref{fig:memory_ablation} (ICEWS14) to ICEWS18, ICEWS05-15, and GDELT.
In each figure, $\Delta_{RR}$ is stratified by 1) train-time history depth (left) and 2) test-time online updates (right).
Across all four benchmarks, every AdaTKG variant lies above the Base in every bin and the gap widens monotonically with the history depth, indicating that the per-entity memory contributes consistently across datasets.

\vspace{25pt}
\section{Gate Distribution Across Benchmarks}
\label{app:gate_violin_grid}

Figure~\ref{fig:gate_violin_all} extends the main-paper Figure~\ref{fig:gate} by showing the full train-time gate distribution stratified by the number of observed interactions, for every (\textit{dataset}, \textit{update operator}) pair.
The same monotonic upward shift in the gate value with more interactions holds across all four benchmarks
and all three update operators,
confirming that the adaptive gate consistently learns to lean on the per-entity memory branch as evidence accumulates, rather than being an artifact of a particular dataset or operator.

\medskip
\noindent\begin{minipage}{\linewidth}
\vspace{15pt}
\centering
\begin{adjustbox}{max width=\linewidth}
\includegraphics[width=\linewidth]{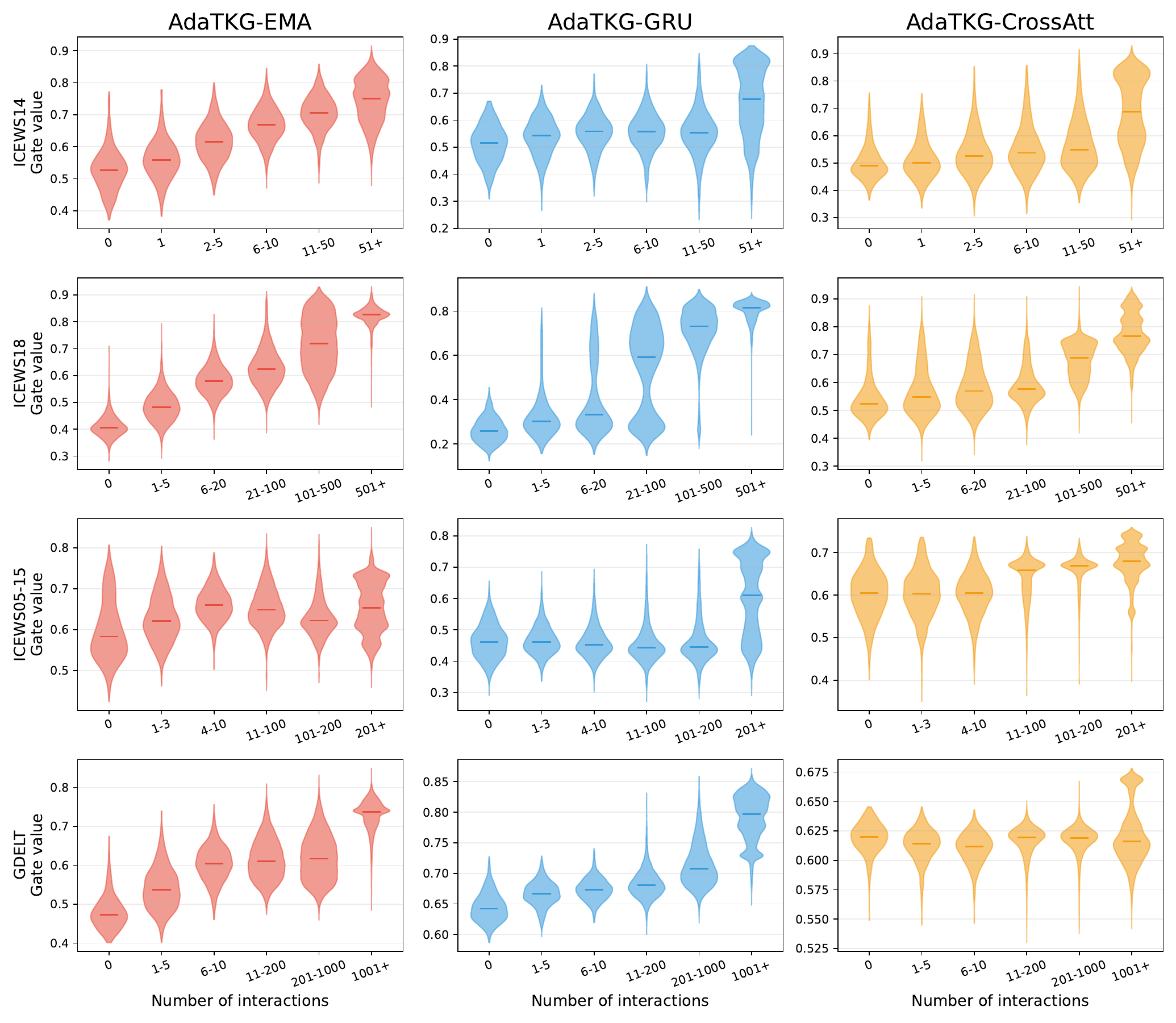}
\end{adjustbox}
\captionof{figure}{\textbf{Gate value by \# interactions.}
Each subplot is a (\textit{dataset}, \textit{operator}) pair, with rows as datasets and columns as AdaTKG operators (EMA, GRU, CrossAtt). Within each subplot, violins show the train-time distribution of the learned gate $g^{(t_q)}_e$ stratified by the number of subject interactions observed before the query, with the median marked.
}
\label{fig:gate_violin_all}
\end{minipage}

\end{document}